\documentclass[sigconf]{acmart}

\copyrightyear{2026}
\acmYear{2026}
\setcopyright{cc}
\setcctype{by}
\acmConference[KDD '26]{Proceedings of the 32nd ACM SIGKDD Conference on Knowledge Discovery and Data Mining V.2}{August 09--13, 2026}{Jeju Island, Republic of Korea}
\acmBooktitle{Proceedings of the 32nd ACM SIGKDD Conference on Knowledge Discovery and Data Mining V.2 (KDD '26), August 09--13, 2026, Jeju Island, Republic of Korea}
\acmDOI{10.1145/3770855.3818090} \acmISBN{979-8-4007-2259-2/2026/08}
\usepackage[utf8]{inputenc}
\usepackage{graphicx}
\usepackage{tabularx}
\usepackage{makecell}
\usepackage{verbatim}
\usepackage{pdfpages}
\usepackage{amsmath, array}
\usepackage{physics}
\usepackage{booktabs}       
\usepackage{multirow}
\usepackage{amsfonts}       
\usepackage{nicefrac}       
\usepackage{microtype}      
\usepackage{balance}
\usepackage{overpic}
\usepackage{subcaption}
\usepackage{xspace}
 \usepackage{xcolor}
\usepackage{pifont}
\usepackage{colortbl}

\definecolor{lightgray}{RGB}{240,240,240}
\definecolor{darkgray}{RGB}{220,220,220}
\newcommand{\largeRedCross}{{\textcolor{red}{\ding{55}}}}
\newcommand{\largeGreenCheck}{{\textcolor{green}{\ding{51}}}}

\newcommand{\ALPHAHILL}{PL\_\allowbreak Alpha\_\allowbreak Hill\xspace}

\newcommand{\LINEARMAP}

\definecolor{simHigh}{RGB}{226,246,226}
\definecolor{simMid}{RGB}{225,230,255}   
\definecolor{simLow}{RGB}{255,232,232}
\usepackage{pgf}
\newcommand{\simcell}[1]{%
  \begingroup
  \pgfmathparse{#1}\let\raw=\pgfmathresult  \pgfmathparse{floor(\raw*100)/100}\let\trunc=\pgfmathresult
  \pgfmathprintnumberto[fixed,precision=2,fixed zerofill]{\trunc}{\disp}%
  \ifdim\raw pt>0.8pt
    \cellcolor{simHigh}\disp%
  \else\ifdim\raw pt>0.5pt
    \cellcolor{simMid}\disp%
  \else
    \cellcolor{simLow}\disp%
  \fi\fi
  \endgroup
}

\newcommand{\legendbox}[1]{%
  \begingroup
  \setlength{\fboxsep}{1.2pt}%
  \colorbox{#1}{\strut\hspace{1.2em}}%
  \endgroup
}

\begin{document}

\title{Spectral Signatures of Large Language Models}

\author{Zhuoying Zhang}
\authornote{Both authors contribute equally to this research.}
\email{zz183@rice.edu}
\affiliation{%
  \institution{Rice University}
  \city{Houston}
  \state{TX}
  \country{USA}
}

\author{Ishan V. Prasad}
\authornotemark[1]
\email{ishan.v.prasad.26@dartmouth.edu}
\affiliation{%
  \institution{Dartmouth College}
  \city{Hanover}
  \state{NH}
  \country{USA}
}

\author{Yuanzhe Hu}
\email{yzhu.ml@outlook.com}
\affiliation{%
  \institution{Georgia Institute of Technology}
  \city{Atlanta}
  \state{GA}
  \country{USA}
}

\author{Zihang Liu}
\email{zihang.liu@berkeley.edu}
\affiliation{%
  \institution{University of California, Berkeley}
  \city{Berkeley}
  \state{CA}
  \country{USA}
}

\author{Hengrui Luo}
\email{hl180@rice.edu}
\affiliation{%
  \institution{Rice University}
  \city{Houston}
  \state{TX}
  \country{USA}
}

\author{Pu Ren}
\authornote{Corresponding author.}
\email{pren@lbl.gov}
\affiliation{%
  \institution{Lawrence Berkeley National Laboratory}
  \city{Berkeley}
  \state{CA}
  \country{USA}
}

\author{Yaoqing Yang}
\email{yaoqing.yang@dartmouth.edu}
\affiliation{%
  \institution{Dartmouth College}
  \city{Hanover}
  \state{NH}
  \country{USA}
}

\renewcommand{\shortauthors}{Zhuoying Zhang et al.}

\begin{abstract}

The rapidly growing repository of publicly available large language models (LLMs) presents significant challenges for systematic management and quantification at scale, such as model lineage tracing, licensing, and evaluation. However, task-specific benchmarks are insufficient for this setting, as LLMs differ widely in architectures, scales, and training procedures. To address this challenge, we adopt \emph{spectral shape}-based metrics for managing and quantifying LLMs based on Heavy-Tailed Self-Regularization theory. Our approach uses the shape information of the weight empirical spectral density as a compact spectral signature of each model. This signature captures intrinsic properties of pretrained models and remains robust during post-training, making it suitable for model-level analysis. In addition, this metric is \emph{data-free}, \emph{computationally-efficient}, and \emph{scale-invariant}, enabling large-scale analysis in practice. Moreover, we curate a large and diverse model corpus consisting of major open-source LLM families, and use it to systematically benchmark spectral and non-spectral metrics across models and downstream tasks. We show that our spectral signature supports the tracking of the model lineage, the unsupervised clustering of similar models, and the quantification of the model performance. Overall, the proposed spectral signature provides a meaningful proxy for broad performance trends across LLMs, enabling efficient organization, comparison, and analysis of large model collections. The code is available at \url{https://github.com/Ingrid-505/Spectral_Signature}.

\end{abstract}


\maketitle

\section{Introduction}
\label{sec:intro}

Publicly available large language models (LLMs) have been growing rapidly in recent years, with hundreds of thousands of models now available through open-source platforms such as Hugging Face (HF)~\citep{zhuang2025embedllm}. These models exhibit significant diversity in their training trajectories, including variations in model architecture, parameter scale, training data, and empirical performance~\citep{horwitz2025Atlas}. As the ecosystem grows, there is an increasing need for reliable and efficient methods to characterize, compare, and manage LLMs beyond task-level benchmarks. Representative examples include identifying whether deployed LLM systems can be traced to the same backbone foundation model for intellectual property and safety purposes~\citep{zhu2025independence}, as well as tracking and organizing model variants and versions in large-scale LLM databases~\citep{horwitz2025TreeHeritage}. Moreover, LLMs present distinct idiosyncrasies across downstream tasks, with different models specializing in different areas such as code generation, mathematical reasoning, long-context processing, or interactive dialogue. These considerations point to the need for a \emph{principled} and \emph{data-light} framework to trace model sources, capture model idiosyncrasies, and quantify model performances for large-scale LLM collections.

In this work, we introduce a \emph{spectral shape}-based perspective for LLM quantification that builds on insights from Heavy-Tailed \- Self-Regularization (HT-SR) theory~\citep{martin2021implicit,martin2020predicting_NatComm,martin2025setol}. HT-SR theory analyzes the eigenspectrum of weight matrices, and prior studies~\cite{yang2023test,martin2020predicting_NatComm} have shown that the shape of the empirical spectral density (ESD) encodes informative signals about model capacity, quality, and training dynamics that are not well captured by scale-based measures. Moreover, ESD shape metrics are data-independent, computationally lightweight, and remain robust under noise and scaling. These properties make them particularly well-suited to large-scale LLM settings, where task-specific evaluation can be expensive. An overview of different quantification metrics is shown in Tab.~\ref{table:metric_overview}.

We use the shape metric of the ESD, namely \ALPHAHILL, as a \emph{spectral signature} of each LLM. The \ALPHAHILL~metric summarizes the global structure of layer-wise weight eigenspectra into a compact representation, enabling efficient storage and comparison across large model collections. We adopt this spectral signature to support three downstream tasks: supervised model classification, unsupervised clustering of LLMs, and quantifying LLM performance.
Our contributions are summarized below.
\begin{itemize}
    \item We introduce a shape-based metric for reliably and efficiently characterizing, comparing, and quantifying LLMs beyond task-level benchmarks. Our metric is \emph{scale-invariant}, \emph{data-free}, \emph{theory-driven}, and \emph{easy-to-compute}. 
    \item We curate a diverse and representative LLM corpus of up to 499 models and use it to benchmark three tasks: LLM similarity measurement, model classification and clustering, and performance prediction. We further conduct a systematic comparison of our spectral signature against a range of baselines across various experimental settings, demonstrating \emph{consistent effectiveness} across tasks and conditions.
    \item We compare shape- and scale-based metrics under output-invariant reparameterizations and noise perturbations. ESD shape metrics remain stable, while scale-based baselines degrade, indicating that spectral shape captures intrinsic weight structure for reliable LLM quantification.
\end{itemize}

\begin{table*}[t!]
  \caption{Comparisons of different metrics for LLM tasks. \largeGreenCheck~and \largeRedCross~denote strong and weak performance, respectively.}
  \label{table:metric_overview}
  \centering
  \resizebox{0.98\linewidth}{!}{
  \begin{tabular}{lcccccccc}
    \toprule 
    \textbf{Metrics} & \textbf{Scale-invariant} & \textbf{Data-free} & \textbf{Easy to compute} & \textbf{Scalable} & \textbf{Theory-driven} & \textbf{Classification} & \textbf{Clustering} & \textbf{Quantification}  \\ 
    \midrule
    
    Logits~\cite{Yang2024AFF} &  
    \largeGreenCheck &  \largeRedCross & \largeGreenCheck & \largeRedCross & \largeRedCross & \largeGreenCheck & \largeGreenCheck & \largeRedCross \\
    \midrule

    REEF~\cite{zhang2025reef} &  
    \largeGreenCheck &  \largeRedCross & \largeGreenCheck & \largeRedCross & \largeRedCross & \largeGreenCheck & \largeGreenCheck & \largeRedCross\\
    \midrule
    
    PCS~\cite{zeng2024HuRef} &  
    \largeRedCross &  \largeGreenCheck & \largeGreenCheck & \largeRedCross & \largeRedCross & \largeGreenCheck & \largeGreenCheck & \largeRedCross \\
    \midrule

    Spectral norm &  
    \largeRedCross &  \largeGreenCheck & \largeGreenCheck & \largeGreenCheck & \largeGreenCheck & \largeGreenCheck & \largeGreenCheck & \largeGreenCheck\\
    \midrule
    
    Log alpha norm &  
    \largeRedCross &  \largeGreenCheck & \largeGreenCheck & \largeGreenCheck & \largeGreenCheck & \largeGreenCheck & \largeGreenCheck & \largeGreenCheck \\
    \midrule
    
    \textbf{Shape metric (Ours)} & \largeGreenCheck  &  \largeGreenCheck & \largeGreenCheck & \largeGreenCheck & \largeGreenCheck & \largeGreenCheck & \largeGreenCheck & \largeGreenCheck \\
    \bottomrule
  \end{tabular}}
\end{table*}

\section{Related Work}
\label{s:related_work}

\noindent\paragraph{Model Lineage and Independence Testing.}
The growing population of open-source LLMs has spurred research into determining LLM lineage—whether two models are independently trained or one is trained from another—which is crucial for intellectual property and model safety. Recent works use distinct approaches: output-based methods use output comparison~\citep{Nikolic2025model}, fingerprint queries~\citep{jin2024proflingo}, and lexical cues~\citep{sun2025idiosyncrasies, wu-etal-2023-llmdet} to compare models without requiring parameter access; watermarking techniques inject
provenance information during training or decoding~\citep{wang2019attacks, Uchida2017embedding, Gu2022WatermarkingPL, jiang2025stealthink}; whereas white-box methods perform hypothesis testing~\citep{zhu2025independence} or fingerprinting using metrics derived from model weight structures~\citep{zeng2024HuRef, zeng2025awm, wang2025ghost, yoon2025intrinsic}.

\paragraph{Model Zoos and LLM Embeddings.} 
There has been growing interest in exploring representation-based approaches to capture model dependence, properties, and relationships, enabling efficient organization and management of large model zoos. Relevant directions include LLM embeddings~\citep{zhuang2025embedllm}, weight-space learning~\cite{horwitz2025Atlas, horwitz2025TreeHeritage, gueta2023KnowledgeRegion}, and loss-landscape-based grouping~\cite{gueta2023KnowledgeRegion, schurholt2025ModelZoo}. LLM embedding methods~\citep{zhuang2025embedllm} learn compact and model-level representations that can be used to predict downstream performance across tasks without additional inference. In contrast, weight-space learning approaches operate directly on model parameters. For example, Horwitz \emph{et al.}~\cite{horwitz2025Atlas} combine weight-space distances with model metadata to construct an ``atlas’’ of open models, enabling similarity search and provenance analysis within large repositories. A complementary line of work~\citep{schurholt2025ModelZoo} considers loss-landscape and phase-based information~\citep{Yang2021TaxonomizingLossLandscapes}, such as mode connectivity and CKA similarity, to partition models into distinct regimes and assess their similarity.

\noindent\paragraph{HT-SR Theory.} HT-SR theory is a theoretical framework based on random matrix theory for analyzing weight matrices geometry of neural networks (NNs)~\citep{mahoney2019traditional, martin2021implicit, couillet2022random, liao2025random}. HT-SR theory was proposed based on the observation that well-trained, state-of-the-art NNs often exhibit HT structures in the ESD of each layer. In the meantime, several rigorous theories in stochastic gradient descent relating HT phenomena to generalization performance were established, providing further theoretical support for HT-SR theory~\citep{gurbuzbalaban2021heavy, hodgkinson2021multiplicative, hodgkinson2022generalization, simsekli2019tail, simsekli2020hausdorff}. Recent works extend theoretical foundations and show that HT-SR Theory can be used to assess the training quality of modern deep NN models across various domains (e.g., computer vision and natural language processing) without accessing any training or test data~\citep{yang2023test, martin2021implicit}. Based on this insight, recent studies have further proposed HT-SR theory-based methods for practical NN applications, including hyperparameter tuning~\citep{zhou2023temperature, liu-etal-2024-model, hu2025eigenspectrum} and model pruning~\citep{lu2024alphapruning,hu2025eigenspectrum}. In this work, we use the HT-SR theory to analyze the geometric similarities of weight matrices of LLMs.

\section{Preliminaries}
\label{sec:preliminary}

\paragraph{Problem Setup.}

In large LLM repositories, many checkpoints share a common pretrained backbone but differ through post-training, distillation, pruning/quantization, or model merging. Our goal is to perform \emph{data-free} lineage analysis: comparing models and tracing backbone relationships directly from their weights. This requires robustness to output-invariant transformations, such as uniform scaling, hidden-unit permutations, and linear-chain reparameterizations, which can make naive weight-space similarities unreliable. We therefore focus on spectral shape statistics of weight matrices, which capture intrinsic weight structures while being less sensitive to transformations that change the parameterization but preserve the model's input-output function.

\paragraph{Weight Analysis with HT-SR Theory.}
HT-SR theory~\citep{martin2021implicit} demonstrates the empirical fact that well-trained NNs tend to exhibit strong correlations in weight matrices, resulting in a heavy-tailed ESD of each weight matrix. ESD characterizes the eigenvalue distribution of the correlation matrix $\mathbf{W}^{\top}\mathbf{W}$ of the weight matrix $\mathbf{W}$. HT-SR theory can be used to analyze the dynamics of weight matrix ESDs during training, by modeling the tail component of the ESD with a power-law (PL) distribution within the interval $(\lambda_{min}, \lambda_{max})$, and its tail index $\alpha$: $p(\lambda) \propto \lambda^{-\alpha}, \quad \lambda_{min} < \lambda < \lambda_{max}$.

One commonly used method to approximate the tail index $\alpha$ is with the Hill Estimator\footnote{It is known that more mathematically principled estimators exist, such as those based on MLE~\cite{alstott2014powerlaw,clauset2009power} or eigenvalue repulsion~\cite{hodgkinson2025models}. However, in our experiments, the Hill estimator demonstrates more stable performance and is easier to scale to large settings.}~\citep{hill1975simple}, which computes an estimate \ALPHAHILL:
\begin{equation}
    \text{\ALPHAHILL} = 1 + \frac{k}{\sum_{i=1}^{k}\operatorname{ln}\frac{\lambda_{n-i+1}}{\lambda_{n-k}}},
\end{equation}
where $\{\lambda_i\}_{i=1}^n$ is sorted in ascending order. For the selection of $k$, we adopt the approach from previous works~\citep{zhou2023temperature, yang2023test}, where we set $k=\frac{n}{2}$. \ALPHAHILL measures the shape of the eigenspectrum of each layer individually. Unlike norm-based metrics such as the spectral norm or RMS norm, \ALPHAHILL is robust under scaling. Moreover, since permutations are orthogonal changes of basis, the spectrum of $\mathbf{W}^{\top}\mathbf{W}$ (and hence its ESD shape) is invariant to hidden-unit permutations, motivating \ALPHAHILL as a stable building block for our spectral signatures (Sec.~\ref{sec:signature_formulation}).
\ALPHAHILL also measures the dynamics of the ESD shape, and previous works~\citep{yang2023test, liu-etal-2024-model} have shown that \ALPHAHILL reflects the training quality of different layers. Layers with larger \ALPHAHILL tend to be ``over-trained,'' while layers with smaller \ALPHAHILL tend to be relatively ``under-trained.''

\begin{figure}[tbh]
    \centering
    \includegraphics[width=0.95\linewidth]{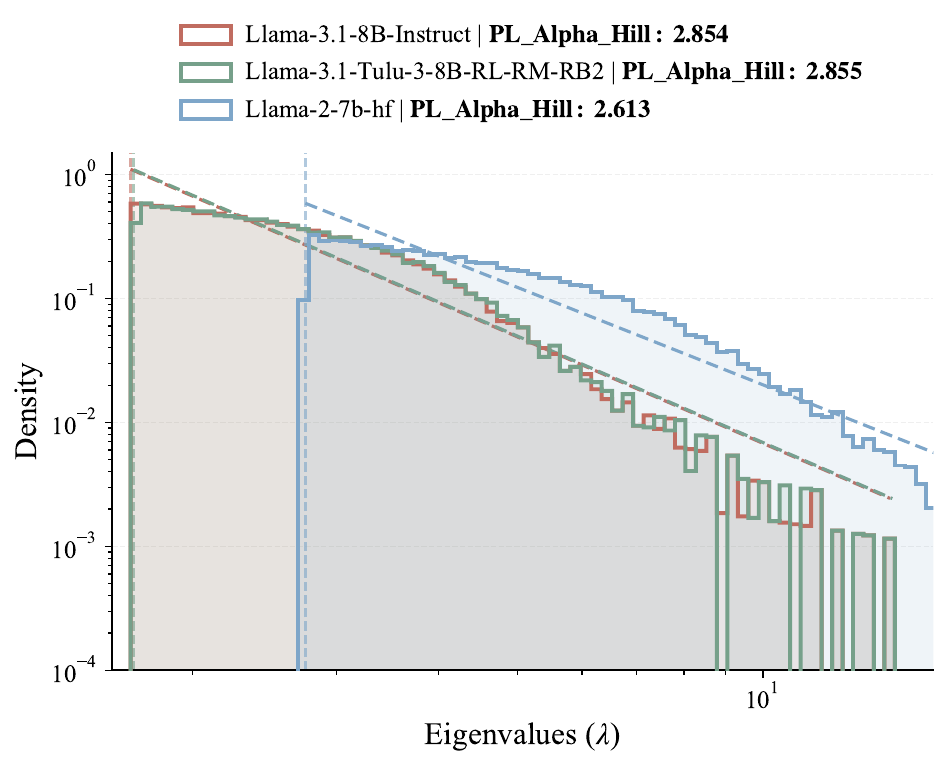}
    \caption{Eigenspectrum shape serves as a distinct signature for model lineage. We visualize the weight ESDs of three LLMs. Models sharing the same backbone (Llama-3.1 and its post-trained variant Tulu-3) exhibit nearly identical spectral shapes and power-law exponents ($\ALPHAHILL \approx 2.85$), while a model from a different family (Llama-2) shows significant deviation. This structural persistence validates the use of ESD shape as a robust spectral signature.}
    \label{fig:esd_layer10_llama}
\end{figure}

\paragraph{Robustness of ESD Shape Metrics in Post-training.}
A critical prerequisite for a spectral signature to serve as a reliable lineage identifier is its robustness across downstream adaptations. We validate the robustness of the layer-wise \ALPHAHILL as a spectral signature by examining its stability under post-training transformations. We observe that the impact of post-training on the weight ESD is minimal; the spectral signature is predominantly shaped during the pre-training phase. As shown in Fig. ~\ref{fig:esd_layer10_llama}, the spectral signature exhibits distinct behaviors for intra-family and inter-family comparisons. We observe a near-perfect spectral overlap between Llama-3.1 and its post-trained Tulu-3 variant. Simultaneously, the separation from the Llama-2 baseline is clear. This distinction supports that the spectral signature is primarily determined during pre-training and remains robust to post-training adjustments. We provide additional results in Appendix~\ref{app:robust_esd} to further corroborate these findings.

\section{Method}
\label{sec:method}

\begin{figure*}[!ht]
\centering
\includegraphics[width=0.98\textwidth]{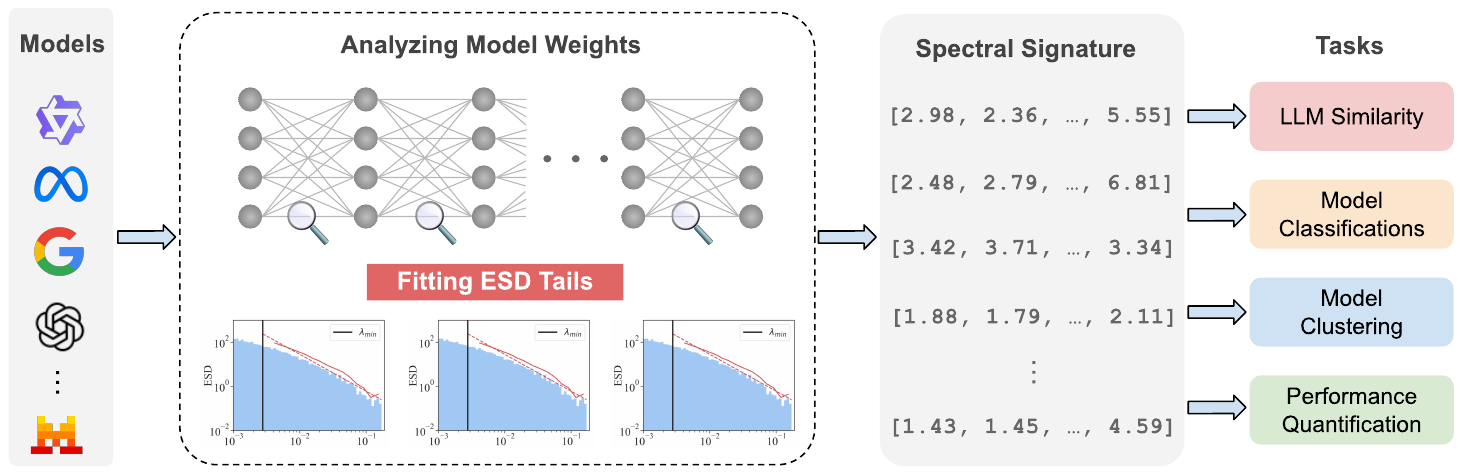}
\caption{Overview of our proposed shape-based spectral signature framework. We analyze the eigenspectrum of model weights and use the shape information of ESDs (\ALPHAHILL) as spectral signatures for downstream tasks.}
\label{fig:teaser}
\end{figure*}

Using robustness properties of \ALPHAHILL established in Sec.~\ref{sec:preliminary}, we present our methodology for applying \ALPHAHILL-based spectral signatures to model quantification. As illustrated in Fig.~\ref{fig:teaser}, for a given collection of LLMs, we extract the \ALPHAHILL statistic from each weight matrix and aggregate these measurements to construct a model-level spectral signature, as formalized in Sec.~\ref{sec:signature_formulation}. We then use these spectral signatures to address three model quantification tasks: (1) measuring similarity between LLMs, (2) model classification and clustering, and (3) predicting downstream LLM performance. The detailed methodology is shown in Sec.~\ref{sec:applying_spectral_signature}.

\subsection{Formulation of Spectral Signatures}
\label{sec:signature_formulation}

For a model $f_{\theta}$ in parameter space $\Theta$ with $L$ transformer blocks and $N_{\text{mod}}$ modules, we quantify the model by constructing a spectral signature: $\mathbf{Z}\in\mathbb{R}^{N_{\text{mod}}\times L}$, 
which is a tensor consisting of the \ALPHAHILL measure of each module. Each row $\mathbf{Z}_{i, :}\in\mathbb{R}^{L}$ is a vector of \ALPHAHILL of modules of the same type. For a standard transformer architecture, we have $N_{\text{mod}} = 7$. 
The tensor representation of spectral signature creates a mapping $\Theta \to \mathbb{R}^{N_{\text{mod}}\times L}$ that extracts spectral information of each weight matrix in the model. It allows module-level analysis, as different types of transformer layers often undergo different degrees of structural changes during post-training. Our empirical observation shows that Multi-Layer Perceptron (MLP) projection layers present larger separation in independently trained models compared to attention layers.

\subsection{Spectral Signatures for Model Quantification}
\label{sec:applying_spectral_signature}

The spectral signature encodes robust spectral information of each weight matrix in an LLM, which can be used for systematic management and quantification of large collections of LLMs. In this section, we describe how spectral signatures can be systematically applied to model quantification tasks, and we demonstrate their use in the following representative scenarios.

\paragraph{Measuring LLM Similarities.}

To quantify the relations between models in a model collection $\mathcal{F}$, we define a similarity measure based on their spectral signatures. Let $\mathbf{Z}^{(k)} \in \mathbb{R}^{N_{\text{mod}} \times L}$ denote the spectral signature of model $k$, where the $i$-th row $\mathbf{Z}^{(k)}_{i, :}\in\mathbb{R}^{L}$ corresponds to the vector of \ALPHAHILL of all occurrences of the $i$-th module type (e.g., Attention Q/K/V, MLP). Given two models $f_{\theta_1}, f_{\theta_2}$ with signatures $\mathbf{Z}^{(1)}, \mathbf{Z}^{(2)}$, we compute the pairwise similarity by averaging the Spearman rank correlations across aligned modules:
\begin{equation}
\label{eq:similarity}
    \mathrm{Sim}(\mathbf{Z}^{(1)}, \mathbf{Z}^{(2)}) = \frac{1}{N_{\text{mod}}} \sum_{i=1}^{N_{\text{mod}}} \rho_s(\mathbf{Z}^{(1)}_{i, :}, \mathbf{Z}^{(2)}_{i, :}),
\end{equation}
where $\rho_s(\cdot, \cdot)$ is the Spearman's rank correlation coefficient:
\begin{equation}
\label{eq:sr}
    \rho_s(\mathbf{u}, \mathbf{v}) = \mathrm{corr}\!\left(\mathrm{rank}(\mathbf{u}),~\mathrm{rank}(\mathbf{v})\right).
\end{equation}
We use Spearman correlation in Eqs.~\ref{eq:similarity}--\ref{eq:sr} because it depends only on the relative depth-wise ordering of \ALPHAHILL values, making the similarity less sensitive to module-specific affine distortions of the statistics and improving comparability across scales.
More importantly, \ALPHAHILL itself provides exact invariances that directly match our output-invariant stress tests.
For any weight matrix $\mathbf{W}$, let $\mathbf{C}=\mathbf{W}^{\top}\mathbf{W}$ with eigenvalues $\{\lambda_j\}$.
If $\mathbf{W}' = c\,\mathbf{P}\mathbf{W}\mathbf{Q}$ where $c>0$ and $\mathbf{P},\mathbf{Q}$ are orthogonal (including permutations), then ${\mathbf{W}'}^{\!\top}\mathbf{W}' = c^2\,\mathbf{Q}^{\top}\mathbf{C}\mathbf{Q}$ has eigenvalues $\lambda'_j=c^2\lambda_j$.
Since the Hill estimator used for \ALPHAHILL depends on log-ratios of tail eigenvalues, these ratios are unchanged and thus $\ALPHAHILL(\mathbf{W}')=\ALPHAHILL(\mathbf{W})$.
This yields permutation- and scale-robust similarity without solving explicit neuron-matching problems that arise in matrix-level fingerprints such as \textsc{AWM}~\cite{zeng2025awm}.

\paragraph{Model Classification}
The properties of the spectral signature further enable its use for identifying the lineage of an LLM among candidate base models, as models derived from the same base model exhibit highly similar eigenspectrum shapes, whereas models with different origins show substantial divergence in shape. We approach the lineage identification as an unsupervised classification problem in the spectral signature space, and we use a nearest-neighbor classifier based on the spectral similarity from Eq.~\ref{eq:similarity}.
Let $\{\mathbf{Z}^{(k)}\}_{k=1}^{K}$ denote the set of spectral signatures for $K$ known base foundation models. For a given query model with spectral signature $Z$, we compute its similarity to each base model via the spectral signature similarity formula defined in Eq.~\ref{eq:similarity}. The predicted lineage $\hat{k}$ is the base model that maximizes this similarity:
\begin{equation}
\label{eq:classification}
\hat{k} = \operatorname*{arg\,max}_{k \in \{1, \dots, K\}} \operatorname{Sim}(\mathbf{Z}, \mathbf{Z}^{(k)})
\end{equation}

\paragraph{Model Clustering.}
Spectral signature of LLMs further enables the discovery of underlying genealogical structures within large model collections without supervision. We formulate this problem as a clustering problem in the space of spectral signatures. Given a model collection $\mathcal{F} = \{f_{\theta_1}, \dots, f_{\theta_N}\}$, we extract the spectral signature $\mathbf{Z}^{(k)} \in \mathbb{R}^{N_{\text{mod}} \times L}$ of each model $f_{\theta_k}$. For vector-based clustering algorithms, we flatten $\mathbf{Z}^{(k)}$ into a feature vector $\mathbf{z}^{(k)} \in \mathbb{R}^{N_{\text{mod}} \cdot L}$. We use spectral signatures for four unsupervised pipelines: HDBSCAN, Bayesian Gaussian Mixture Models (BGMM), Spectral Clustering, and K-Means. Due to the high dimensionality of the spectral signature representations, we apply Uniform Manifold Approximation and Projection (UMAP) as a preprocessing step for HDBSCAN, BGMM, and K-Means. In contrast, Spectral Clustering operates directly on the original high-dimensional pairwise distance matrix, where dimensionality reduction is implicitly handled through the graph Laplacian eigenstructure. Hyperparameters for clustering algorithms (i.e., number of clusters) are selected by maximizing the silhouette score. To ensure that clustering quality reflects intrinsic structure in the weight space rather than artifacts introduced by the embedding process, all internal evaluation metrics (e.g., silhouette scores) are computed using distances in the original spectral signature space. We assess the robustness and generalizability of each clustering pipeline by repeating experiments under random subsampling at varying sample sizes, verifying consistent clustering behavior across regimes.

\paragraph{Predicting LLM Performance.}
To quantify the predictive power of the proposed spectral signature, we treat the performance estimation as a non-parametric regression task over the model manifold. Given a query model $f_{\theta}$ with spectral signature $\mathbf{Z}$ and a reference pool of models $\mathcal{P} = \{(\mathbf{Z}^{(i)}, \mathbf{b}_i)\}_{i=1}^N$, where $\mathbf{Z}^{(i)}$ is the spectral signature and $\mathbf{b}_i$ is the vector of benchmark scores, we estimate the performance of $f_{\theta}$ using distance-weighted $k$-nearest neighbor ($k$-NN) Interpolation.
Specifically, we first derive a distance metric $d(\mathbf{Z}, \mathbf{Z}^{(i)})$ from the spectral similarity $\mathrm{Sim}(\mathbf{Z}, \mathbf{Z}^{(i)})$. 
We employ the transformation $d(\mathbf{Z}, \mathbf{Z}^{(i)})= 1 - \mathrm{Sim}(\mathbf{Z}, \mathbf{Z}^{(i)})$. This ensures that models with highly correlated spectral shapes are positioned closer in the embedding space. The predicted benchmark scores $\tilde{\mathbf{b}}$ are computed by aggregating the top-$k$ most similar neighbors in $\mathcal{P}$:
$$\tilde{\mathbf{b}} = \frac{\sum_{i \in \mathcal{K}} w_i \cdot \mathbf{b}_i}{\sum_{i \in \mathcal{K}} w_i},$$,
where $\mathcal{K}$ is the set of $k$ nearest neighbors, and $w_i = 1 / (d(\mathbf{Z}, \mathbf{Z}^{(i)}) + \epsilon)$ is the assigned weight. This interpolation relies on the assumption that models that exhibit similar HT-SR patterns in their weight matrices are likely to possess comparable generalization capabilities. We then quantify the prediction accuracy with:
\begin{equation}
D_{perf}=\|\tilde{\mathbf{b}}-\mathbf{b}\|,
\label{eq:Perfdist}
\end{equation}
where $\mathbf{b}$ represents the ground truth benchmark scores. Typically, $\|\cdot\|$ in Eq.~\ref{eq:Perfdist} is defined by mean absolute error (MAE).

\section{Experiments}\label{sec:results}

In this section, we demonstrate that spectral signatures provide an efficient and accurate tool for a range of downstream tasks. Specifically, we present details of the experimental setup as well as the results of (i) measuring LLM similarities, (ii) model classification and clustering, and (iii) predicting and ranking model performance. 

\subsection{Measuring LLM Similarities}\label{sec:result_llm_lineage}

\paragraph{Model Corpus.} 
We select Llama-2-7b~\cite{touvron2023llama2openfoundation} as the base model and compare it with four types of related variants that share the same architecture:
(i) post-trained models (SFT and RL),
(ii) merged models and pruned models,
(iii) noise-perturbed models obtained by adding i.i.d. Gaussian perturbations to all weight entries with noise strength $\gamma \in \{0.05,0.30,1.00\}$, as defined in Appendix~\ref{ablation_app},
(iv) output-invariant transformations on adjacent matrices and output-invariant permutations of hidden units, as defined in Appendix~\ref{ablation_app}. This corpus isolates a key practical goal: a lineage/similarity signal should remain stable under post-training and output-invariant reparameterizations, while being informative under destructive edits (e.g., aggressive pruning). We further validate the generalizability of our spectral signature to Mixture-of-Experts architectures and cross-depth model pairs in Appendix ~\ref{app:scale}

\paragraph{Setup.}
We compare against both \emph{data-aware} and \emph{data-free} baselines. For data-aware methods, we use \textsc{REEF}~\cite{zhang2025reef}, which computes similarity from activation features on a fixed prompt set (300 prompts for TruthfulQA~\cite{lin2022truthfulqa}), and \textsc{Logits}~\cite{Yang2024AFF}, which measures similarity from model outputs on the same prompts. For data-free approaches, we apply \textsc{PCS}~\cite{zeng2024HuRef} that computes cosine similarity between flattened weight tensors under a fixed alignment, and discuss alignment-based weight signature (e.g., \textsc{AWM}~\cite{zeng2025awm}) as a complementary point of comparison: these methods operate at the matrix level and often require per-pair matching/alignment, which makes them less convenient as a drop-in kernel when building dense similarity matrices for zoo-scale clustering and retrieval.

\paragraph{Results.}
Tab.~\ref{tab:similarity_score} summarizes the overall similarity performance of our method compared to baseline models. It demonstrates the robustness of our spectral signature under diverse scenarios, including fine-tuning, pruning, model merging, noise injection, and output-invariant attacks. First, the similarity scores of our method remain consistently high among Llama-2-7b related variants, such as multiple fine-tuned models, with scores in the $[0.97,1.0]$ range across instruction tuning and domain shifts. In contrast, data-aware baselines can fail under prompt- or language-induced behavior changes. For example, Logits drops sharply for the Chinese variants, and REEF degrades for the LoRA-unloaded model, whereas our scores remain stable. Under weight noise injection, our similarity decreases mildly as $\sigma$ increases from $\approx 0.999$ at $\gamma=0.05$ to $\approx 0.979$ at $\gamma=1.00$, outperforming PCS and REEF at the strongest noise perturbation. 
Moreover, our method is robust to hidden-unit permutations and random scaling under output-invariant transformations, while PCS is sensitive to both.
For unstructured pruning, the similarity score using spectral signature decreases gradually at 30–50\% sparsity and drops substantially at 70\% sparsity, indicating that the spectral signature is informative of the structural changes under more aggressive parameter removal. 

Quantifying similarity among large model corpora typically requires constructing a stable and efficient pairwise similarity matrix. However, existing methods often fall short computationally: data-dependent methods require prompts and incur inference costs for each comparison, while pair-specific methods such as \textsc{AWM} incur cumulative overhead when constructing dense similarity matrices over large model collections.
Our approach computes a compact signature once per model and yields a computationally-cheap similarity that remains robust under the varying transformations that break coordinate-dependent weight distances, making it practical for large-scale lineage analysis.

\begin{table*}[!htbp]
\centering
\caption{A comprehensive evaluation of signature-based methods for measuring LLM similarity across diverse settings. Here, \legendbox{simHigh}, \legendbox{simMid}, and \legendbox{simLow} denote similarity scores $>0.8$ (high), between $0.5$ and $0.8$ (moderate), and $<0.5$ (low), respectively.}
\label{tab:similarity_score}

\renewcommand{\arraystretch}{1.15}
\setlength{\tabcolsep}{5pt} 
\resizebox{0.96\textwidth}{!}{%
\begin{tabular}{l*{6}{>{\centering\arraybackslash}m{2.45cm}}}
\toprule\toprule
\rowcolor{black!10}
 & \multicolumn{6}{c}{\textbf{Model Fine-tuning}} \\
 & \makecell{\textbf{llama2-7b-hf-}\\\textbf{instruction-lora}}
 & \makecell{\textbf{vicuna-}\\\textbf{-v1.5}}
 & \makecell{\textbf{Nous-Hermes-}\\\textbf{llama-2-7b}}
 & \makecell{\textbf{chinese-llama}\\\textbf{-2-7b}}
 & \makecell{\textbf{Llama-2-7b-hf}\\\textbf{-instruct-pl}\\\textbf{-lora\_unload}}
 & \makecell{\textbf{chinese-alpaca}\\\textbf{-2-7b}} \\
\midrule
\textbf{PCS}~\cite{zeng2024HuRef}    & \simcell{0.9934} & \simcell{0.9986} & \simcell{0.9998} & \simcell{0.9390} & \simcell{0.9992} & \simcell{0.9521} \\
\textbf{Logits}~\cite{Yang2024AFF}     & \simcell{0.9885} & \simcell{0.7966} & \simcell{0.9920} & \simcell{0.0318} & \simcell{0.9953} & \simcell{0.5631} \\
\textbf{REEF}~\cite{zhang2025reef} & \simcell{0.7704} & \simcell{0.9989} & \simcell{0.9991} & \simcell{0.9974} & \simcell{0.1463} & \simcell{0.9962} \\
\textbf{GhostSpec}~\cite{wang2025ghost}     & \simcell{0.9848} & \simcell{0.9993} & \simcell{0.9999} & \simcell{0.8078} & \simcell{0.8784} & \simcell{0.8314} \\
\textbf{Ours}  & \simcell{0.9989} & \simcell{0.9994} & \simcell{0.9995} & \simcell{0.9728} & \simcell{0.9994} & \simcell{0.9780} \\
 \midrule\midrule
 \rowcolor{black!10}
  & \multicolumn{6}{c}{\textbf{Unrelated Models}} \\
  & \makecell{\textbf{llama-7b}}
  & \makecell{\textbf{llama-3.1-8B}}
  & \makecell{\textbf{Mistral-7B}}
  & \makecell{\textbf{Qwen-1.5-7B}}
  & \makecell{\textbf{open-llama-7b}}
  & \makecell{\textbf{open-llama-7b}\\\textbf{-v2}} \\
 \midrule
 \textbf{PCS}~\cite{zeng2024HuRef}  & \simcell{0.0151} & \simcell{0.1458} & \simcell{0.2749} & \simcell{0.0647} & \simcell{0.0185} & \simcell{0.0149} \\
 \textbf{Logits}~\cite{Yang2024AFF}  & \simcell{0.9900} & \simcell{0} & \simcell{0.7760} & \simcell{0.0004} & \simcell{0.9606} & \simcell{0.9547} \\
 \textbf{REEF}~\cite{zhang2025reef} & \simcell{0.1597} & \simcell{0.1517} & \simcell{0.1335} & \simcell{0.1496} & \simcell{0.1745} & \simcell{0.1865} \\
  \textbf{GhostSpec}~\cite{wang2025ghost} & \simcell{0.9438} & \simcell{0.1905} & \simcell{0.0092} & \simcell{0.8953} & \simcell{0.9122} & \simcell{0.8039} \\
 \textbf{Ours}  & \simcell{0.9275} & \simcell{0.6621} & \simcell{0.7744} & \simcell{0.6882} & \simcell{0.7238} & \simcell{0.6053} \\
\midrule\midrule
\rowcolor{black!10}
 & \multicolumn{3}{c}{\textbf{Unstructured Pruning}} & \multicolumn{3}{c}{\textbf{RL Models}} \\
 & \makecell{\textbf{30\%}}
 & \makecell{\textbf{50\%}}
 & \makecell{\textbf{70\%}}
 & \makecell{\textbf{Archer-Code-}\\\textbf{1.5B (GRPO)}}
 & \makecell{\textbf{
Nemotron-Resear-}\\\textbf{ch-Reasoning}\\\textbf{-Qwen-1.5B}\\\textbf{(Reinforcement++)}}
 & \makecell{\textbf{
Polaris-7B-}\\\textbf{Preview}\\\textbf{(DAPO)}} \\
\midrule
\textbf{PCS}~\cite{zeng2024HuRef}  & \simcell{0.9907} & \simcell{0.9051} & \simcell{0.7807} & \simcell{0.9920} & \simcell{0.9920} & \simcell{1.000} \\
\textbf{Logits}~\cite{Yang2024AFF} & \simcell{1.000} & \simcell{1.000} & \simcell{0.9950} & \simcell{1.000} & \simcell{1.000} & \simcell{1.000} \\
\textbf{REEF}~\cite{zhang2025reef} & \simcell{0.9983} & \simcell{0.9846} & \simcell{0.9818} & \simcell{0.9866} & \simcell{0.9699} & \simcell{0.9490} \\
\textbf{GhostSpec}~\cite{wang2025ghost}     & \simcell{0.9987} & \simcell{0.8968} & \simcell{0.7045} & \simcell{1.0000} & \simcell{1.0000} & \simcell{0.4674} \\
\textbf{Ours} & \simcell{0.8989} & \simcell{0.8903} & \simcell{0.4033 } & \simcell{0.9997} & \simcell{0.9998} & \simcell{0.9997} \\
\midrule\midrule
\rowcolor{black!10}
 & \multicolumn{3}{c}{\textbf{Noise Injection}} & \multicolumn{3}{c}{\textbf{Weight Merging (Evollm-jp)}} \\
 & \makecell{\textbf{$\gamma=0.05$}}
 & \makecell{\textbf{$\gamma=0.30$}}
 & \makecell{\textbf{$\gamma=1.00$}}
 & \makecell{\textbf{shisa-gamma-7b}}
 & \makecell{\textbf{wizardmath-7b-1.1}}
 & \makecell{\textbf{abel-7b-002}} \\
\midrule
\textbf{PCS}~\cite{zeng2024HuRef} & \simcell{0.9988} & \simcell{0.9584} & \simcell{0.7097} & \simcell{0.9993} & \simcell{0.9988} & \simcell{0.9989} \\
\textbf{Logits}~\cite{Yang2024AFF} & \simcell{1.000} & \simcell{1.000} & \simcell{0.9700} & \simcell{0.0000} & \simcell{1.000} & \simcell{1.000} \\
\textbf{REEF}~\cite{zhang2025reef} & \simcell{0.9990} & \simcell{0.9949} & \simcell{0.6932} & \simcell{0.9333} & \simcell{0.9146} & \simcell{0.8956} \\
\textbf{GhostSpec}~\cite{wang2025ghost}     & \simcell{1.000} & \simcell{0.975} & \simcell{0.706} & \simcell{0.2384} & \simcell{0.9961} & \simcell{0.9968} \\
\textbf{Ours} & \simcell{0.9986} & \simcell{0.9868} & \simcell{0.9786} & \simcell{0.8078} & \simcell{0.8784} & \simcell{0.8314}\\
\bottomrule\bottomrule
\end{tabular}
}
\end{table*}

\subsection{Model Classification and Clustering}
\paragraph{Model Corpus.}
We construct a corpus of 128 publicly available models derived from 8 model families: llama-7b~\cite{touvron2023llamaopenefficientfoundation}, llama-2-7b~\cite{touvron2023llama2openfoundation}, llama-3.1-8b~\cite{grattafiori2024llama3herdmodels}, mistral-7b-v0.1, mistral-7b-v0.2~\cite{jiang2023mistral7b}, open-llama-7b, open-llama-7b-v2~\cite{openlm2023openllama}, and qwen-1.5-7b~\cite{bai2023qwentechnicalreport}. 
For each derived model, we assign a training-lineage label corresponding to one of the eight base models, using the model’s publicly documented training provenance. For the classification task, we classify each of the 128 derived models into one of the 8 base families by selecting the base with the highest spectral signature similarity score to the model. For the clustering task, we perform unsupervised clustering on the signatures of all 128 derived models and 8 base models.

\paragraph{Setup.}
For the classification task, we compute the similarity score using three different kernels on spectral signatures: Spearman rank correlation, radial basis function (RBF) kernel, and cosine similarity (detailed in Appendix Sec.~\ref{app:classification_metric}). This allows us to assess the robustness of classification performance under different similarity measures.
For the clustering task, we conduct two complementary studies. First, we assess clustering performance using spectral signatures constructed with different ESD-based metrics, including shape metrics (\ALPHAHILL), scale metrics (Log\_Norm and Spectral\_Norm), and a combined shape–scale metric (Log\_Alpha\_Norm). Second, we evaluate the robustness of clustering by applying four clustering algorithms, as described in Sec.~\ref{sec:method}, to these spectral signatures. We assess clustering quality using the Silhouette Coefficient~\cite{rousseeuw1987silhouettes} and the Davies–Bouldin Index~\cite{davies2009cluster}, as detailed in Appendix~\ref{app:eval_metrics_clustering}.

\begin{figure}[t]
    \centering
    \includegraphics[width=\linewidth]{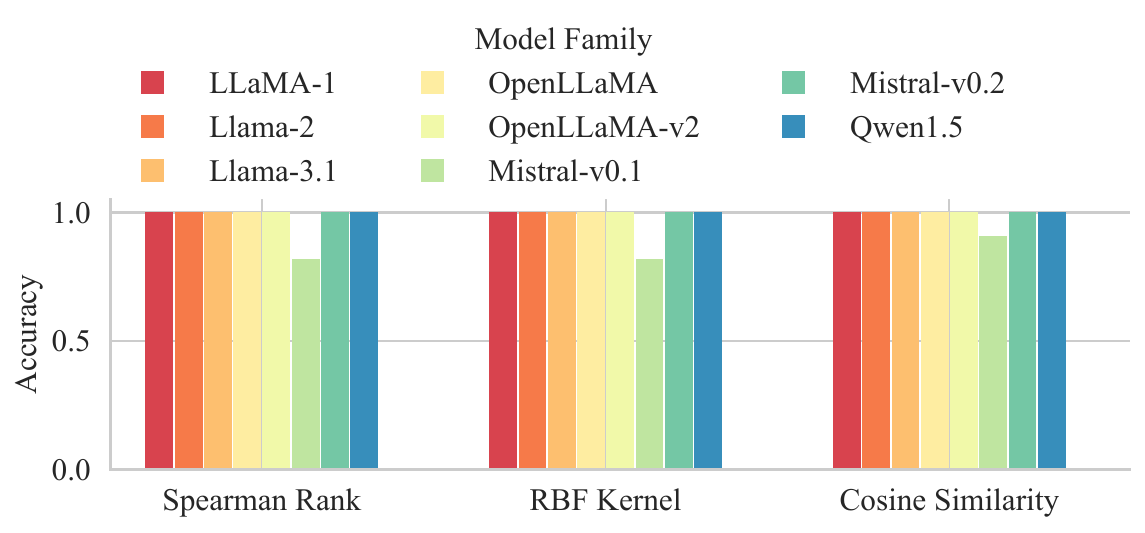}
    \caption{
    Classification accuracy of the 8-way classifier for routing derived models to their base models, reported across model families and similarity kernels. Each bar shows accuracy for one family-kernel pair.
    }
    \label{fig:barchart}
\end{figure}

\paragraph{Classification Results.}
Fig.~\ref{fig:barchart} shows that the proposed spectral signature enables accurate lineage classification across all eight model families with an overall accuracy of 98.44\%. Across similarity kernels, the classifier consistently achieves near-perfect accuracy, indicating that lineage information is strongly preserved in the spectral signature. Performance is stable across Spearman rank, RBF, and cosine similarity, suggesting that the results are not sensitive to the choice of similarity measure. Minor accuracy variations are mainly observed for closely related families, such as mistral-7b-v0.1 and mistral-7b-v0.2, which correspond to different versions within the same model line and share highly similar architectures and training procedures. Overall, these results demonstrate that spectral signatures provide a reliable and discriminative basis for model lineage classification.

\paragraph{Clustering Results.}
Tab.~\ref{tab:clustering_quality_esd} compares the clustering performance of spectral signatures constructed from scale- and shape-based metrics. The proposed shape-based \ALPHAHILL achieves the best performance, with the highest Silhouette score of 0.91 and the lowest Davies-Bouldin index of 0.22, consistently outperforming the scale-based alternatives. To further evaluate the robustness of our proposed spectral signature, in Tab. \ref{tab:clustering_comparison_alpha} we report the result of different clustering algorithms. All clustering algorithms that use the spectral signature achieve high silhouette scores, ranging from 0.73 to 0.91. This shows that the proposed spectral signature provides a robust and effective embedding for different classes of clustering algorithms. As an illustrative example, in Fig.~\ref{fig:tsne} we provide the clustering result of 2D t-SNE using the spectral clustering algorithm with spectral signatures constructed by \ALPHAHILL. As we can see, the proposed spectral signature clearly separates different model families, resulting in distinct clusters.

\begin{table}[h]
\centering
\caption{
Comparison of scale- and shape-based spectral signatures for spectral clustering.
}
\label{tab:clustering_quality_esd}
\resizebox{0.95\columnwidth}{!}{
    \begin{tabular}{lcc}
    \toprule
    \textbf{Metric} & \textbf{Silhouette (↑)} & \textbf{Davies\_Bouldin (↓)}\\
    \midrule
    Log\_Norm (scale)& 0.70 & 0.47\\
    Spectral\_Norm (scale) & 0.81 & 0.35\\
    Log\_Alpha\_Norm (scale) & 0.88 & 0.23\\
    \rowcolor{black!10} \ALPHAHILL (shape) & \textbf{0.91} & \textbf{0.22}\\
    \bottomrule
    \end{tabular}%
}
\end{table}

\begin{table}[h]
\centering
\caption{Comparing different clustering algorithms using the spectral signature constructed by \ALPHAHILL.}
\label{tab:clustering_comparison_alpha}
\resizebox{0.9\columnwidth}{!}{
    \begin{tabular}{lcc}
    \toprule
    \textbf{Method} & \textbf{Silhouette ($\uparrow$)} & \textbf{Davies\_Bouldin ($\downarrow$)}\\
    \midrule
    HDBSCAN & 0.73 & 0.56 \\
    BGMM    & 0.73 & 0.56 \\
    K-Means   & 0.73 & 0.56 \\
    Spectral Clustering & 0.91 & 0.22 \\
    \bottomrule
    \end{tabular}
}
\end{table}

\subsection{Quantifying and Ranking LLM Performance}
\label{sec:results_eval_performance}

\paragraph{Model Corpus.} 
For model ranking and prediction tasks, we curate 499 open-source models from the Open LLM Leaderboard~\cite{myrzakhan2024openllmleaderboard}. This corpus spans a broad spectrum of architectures, focusing on the Llama, Mistral, GPT, and Pythia families, with parameter scales ranging from 19 million to 70 billion. Further validation on additional scales and intra-family prediction is provided in Appendix~\ref{app:scale}.
 
\paragraph{Setup.} 
To evaluate whether spectral signatures can predict downstream performance, we consider four representative benchmarks: ARC-Challenge (reasoning)~\cite{clark2018thinksolvedquestionanswering}, HellaSwag (commonsense)~\cite{zellers-etal-2019-hellaswag}, MMLU (world knowledge)~\cite{hendrycks2021measuring}, and TruthfulQA (factuality)~\cite{lin2022truthfulqa}. 
The models span a wide performance range (e.g., MMLU scores from 23.0 to 70.67), covering both low- and high-performance regimes. For each model, we estimate performance using distance-weighted $k$-NN interpolation ($k=3$) based on spectral-signature similarity. We use 5-fold cross-validation to assess generalization.

We compare against two activation-based baselines. First of all, EmbedLLM~\cite{zhuang2025embedllm} encodes each model by extracting the hidden-state embedding at the final token position from the last transformer layer on a fixed prompt set $\mathcal{P}$, and trains a linear regressor to predict benchmark scores under leave-one-out evaluation. Second, LLMDNA~\cite{wu2025llm} extracts a $128 \times 128$ model representation from 100 random samples, which is nearly invariant to the specific dataset or questions used for extraction. We use the same regression protocol for LLMDNA as for EmbedLLM. Prediction quality is evaluated by MAE and Kendall’s Tau rank correlation~\cite{kendall1948rank}. We define ``Significance'' as the number of repeated train/test splits (out of 100), where Kendall’s Tau between predicted and ground-truth rankings is statistically significant at the 5\% level.

We further discuss why our spectral signatures are well suited for this setting.
Activation-based performance quantification requires task data at inference, and does not scale well to frequent re-ranking across large model collections. Conversely, other weight-based methods such as \textsc{PCS}~\cite{zeng2024HuRef} and \textsc{AWM}~\cite{zeng2025awm} are primarily designed for pair-wise provenance checks and do not naturally provide a compact, geometry-aware representation amenable to regression across heterogeneous architectures. Our spectral signatures provide a lightweight, task-agnostic representation that supports nearest-neighbor interpolation without any prompts.

\paragraph{Results.}

\begin{figure}
    \centering
    \includegraphics[width=0.8\linewidth]{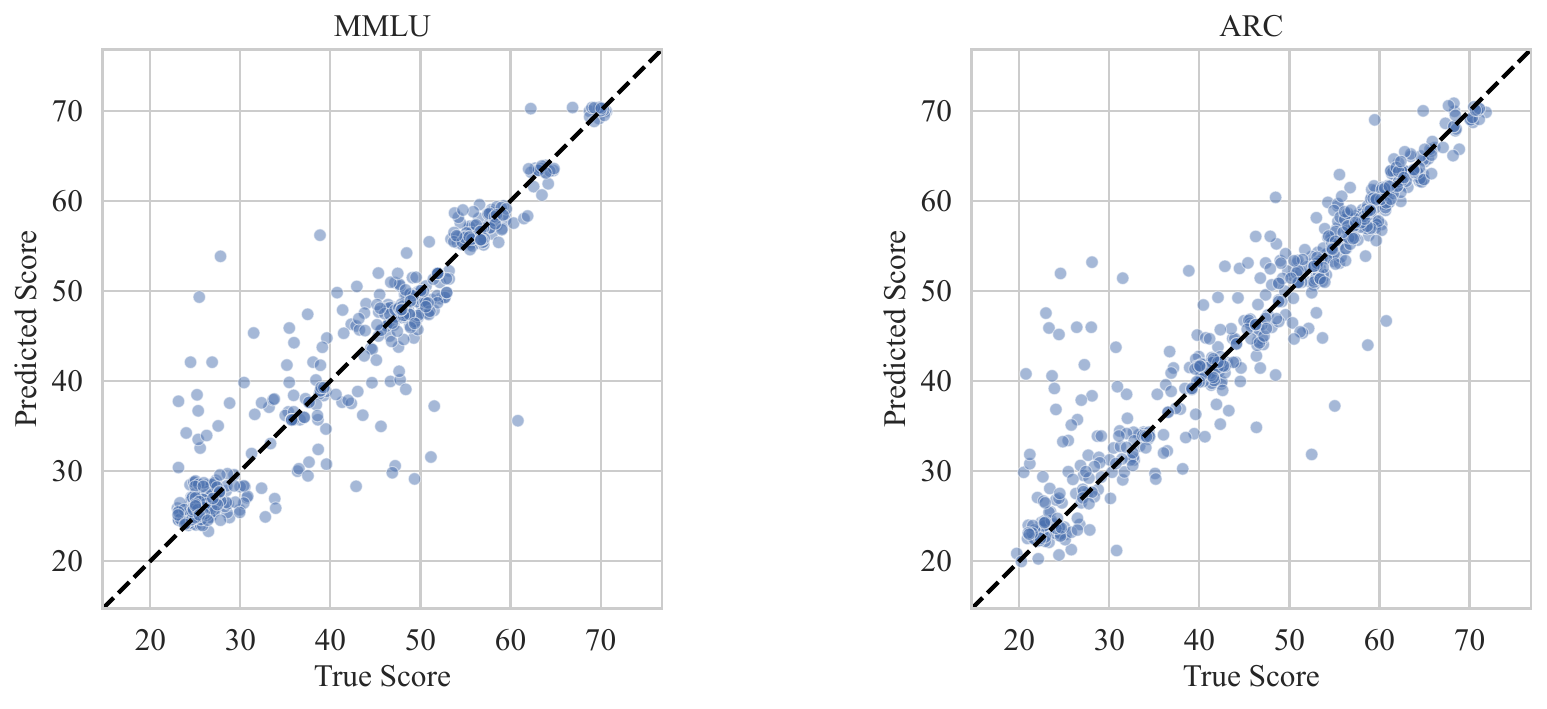}
    \caption{
    Predicted versus ground-truth benchmark scores on MMLU and ARC across the model pool. The dashed diagonal line denotes perfect agreement.
    }
    \label{fig:scatter}
\end{figure}

We run 100 repeated trials with random train/test splits (249 models for training and 250 models for testing in each trial). Fig.~\ref{fig:scatter} compares predicted benchmark scores with their ground-truth values. The points cluster closely around the diagonal ($y=x$), indicating good agreement between predicted and actual performance. 
We report (i) the average MAE over the 100 trials, which measures absolute prediction error, and (ii) the number of trials in which Kendall's Tau correlation between predicted and ground-truth rankings on the test set is statistically significant. As shown in Tab.~\ref{tab:significance}, spectral signatures achieve consistently low MAE and statistically significant rank correlation in all trials across all benchmarks.  While EmbedLLM achieves slightly lower MAE on ARC and TruthfulQA, spectral signatures outperform EmbedLLM on MMLU and remain competitive on HellaSwag. 
This gap reflects that benchmarks targeting specific model capabilities (TruthfulQA, ARC) are more directly reflected in task-specific activations, while benchmarks covering general knowledge (MMLU, HellaSwag) align more closely with the pretraining capacity encoded in weight geometry. 
LLMDNA~\cite{wu2025llm}, despite requiring comparable inference cost, yields substantially higher MAE across all benchmarks. Unlike EmbedLLM, which extracts task-conditioned activations, LLMDNA produces a single task-agnostic embedding per model. The compression loses the benchmark-specific variability needed for accurate performance prediction. 

\paragraph{Computational Cost.}
We report a runtime comparison between our spectral signature and baselines under the same model and hardware settings. We use Llama-2-7b as the representative backbone and run all methods on a single NVIDIA L40 GPU with FP16 precision. For EmbedLLM, we follow the original pipeline and extract last-token hidden-state embeddings on a fixed prompt set. For LLMDNA, we largely follow the original pipeline, with slight modifications to adjust for our hardware and specific models in our dataset.
For our method, we compute the spectral signatures directly from model weights without any forward passes. We measure \textit{Run Time} from the end of model loading to the end of signature/embedding extraction. One iteration of EmbedLLM embeddings takes 2,633 seconds, and one iteration of LLMDNA embeddings takes 5,393 seconds, while the Runtime of Spectral Signature is 337 seconds, significantly more time-efficient than the baselines.

\begin{table}[t]
\centering
\caption{
Performance prediction results on four benchmarks, comparing our spectral signatures, EmbedLLM, and LLMDNA using MAE and statistical significance (Signif.).
}
\label{tab:robustness_knn_embedllm}
\setlength{\tabcolsep}{5pt}
\renewcommand{\arraystretch}{1.2}
\resizebox{\columnwidth}{!}{%
\begin{tabular}{lcccccc}
\toprule
& \multicolumn{2}{c}{\textbf{Ours}} 
& \multicolumn{2}{c}{\textbf{EmbedLLM}} 
& \multicolumn{2}{c}{\textbf{LLMDNA}} \\
\cmidrule(lr){2-3} \cmidrule(lr){4-5} \cmidrule(lr){6-7}
\textbf{Benchmark} & \textbf{MAE} & \textbf{Signif.} & \textbf{MAE} & \textbf{Signif.} & \textbf{MAE} & \textbf{Signif.} \\
\midrule
ARC        & 2.27 & 100 & 1.61 & 100 & 7.12  & 100 \\
HellaSwag  & 3.25 & 100 & 3.50 & 100 & 11.65 & 100 \\
MMLU       & 1.95 & 100 & 3.14 & 100 & 6.78  & 100 \\
TruthfulQA & 2.99 & 100 & 1.96 & 100 & 3.43  & 100 \\
\bottomrule
\end{tabular}}
\label{tab:significance}
\end{table}

\subsection{Generalization to Diverse Architectures }
\label{sec:generalize}

Here, we test our spectral signature on more diverse architectures.

\paragraph{Model Lineage on MoE models}
We evaluate our spectral signature on Pangu-Pro-MoE~\cite{tang2025pangupromoemixture}, an MoE architecture derived from the Qwen2.5-14B family. To construct spectral signatures for MoE modules, we average the spectral metrics across experts. Tab.~\ref{tab:moe} reports the similarity between Pangu-Pro-MoE and candidate source models. Our method identifies Qwen2.5-14B, Qwen2.5-Coder-14B, and OpenMath-Nemotron-14B as the most likely sources, demonstrating that spectral signatures generalize to MoE architectures with minimal adaptation.

\begin{table}[ht!]
    \caption{Similarities between Pangu-Pro-MoE and candidate source models.}
    \label{tab:moe}
    \centering
    \resizebox{0.7\columnwidth}{!}{
        \begin{tabular}{lc}
        \toprule
        \textbf{Candidate Model} & \textbf{Similarity} \\
        \midrule
        Qwen/Qwen2.5-14B & 0.91 \\
        nvidia/OpenMath-Nemotron-14B & 0.87 \\
        Qwen/Qwen2.5-Coder-14B & 0.84 \\
        Qwen/Qwen1.5-14B & 0.30 \\
        01-ai/Yi-1.5-9B & 0.07 \\
        meta-llama/Llama-2-13b-hf & 0.04 \\
        \bottomrule
        \end{tabular}
    }
\end{table}

\paragraph{Similarity Between Depth-Mismatched Models.}
Models derived from the same source may differ in depth. We use an order-preserving layer matching via dynamic programming (DP)\cite{DP} on spectral signatures. We build a cost matrix $D[i,j] = mean|Z_1[i] - Z_2[j]|$ over 7 z-score normalized module columns, then find the injective order-preserving mapping minimizing total cost via DP. We evaluate on the model pair (Llama-3.1-8B vs. Llama-3.2-3B) to enable direct comparison with their activation-based ground truth. Tab.~\ref{tab:match} shows a $3.7\times$ lower matching cost for the related pair. Moreover, the optimal matching skips layers \{11, 17, 23, 29\} of Llama-3.1-8B, consistent with prior activation-based matching results~\cite{zhu2025independence}. This agreement suggests that our data-free spectral signatures capture structural information for cross-depth layer alignment.

\begin{table}[h]
\caption{Matching cost and post-matching Spearman correlation.}
\label{tab:match}
\centering
\small
\begin{tabular}{lcc}
\toprule
\textbf{Candidate Model} & \textbf{Matching cost ($\downarrow$)} & \textbf{Similarity ($\uparrow$)} \\
\midrule
Llama-3.2-3B  & \textbf{0.26} & \textbf{0.89} \\
Qwen2.5-7B    & 0.95          & 0.77 \\
\bottomrule
\end{tabular}
\end{table}
\subsection{Ablation Study}
\label{sec:ablation}

\textbf{\ALPHAHILL remains robust under output-invariant reparameterizations.}
We evaluate whether \ALPHAHILL is a robust metric for constructing a spectral signature by performing output-invariant reparameterizations on weight matrices before measuring model similarity with Eq.~\ref{eq:similarity}. We compare the spectral signature constructed by \ALPHAHILL with other ESD scale metrics (Log Norm and Spectral Norm). Concretely, we apply two reparameterizations on LLaMA-2-7B model: (i) using output-invariant scaling to consecutive attention modules (Q/K and V/O pairs), and (ii) output-invariant change of basis between two consecutive linear layers in the feed-forward network (FFN) module (more details can be found in Appendix~\ref{ablation_app}).
From Fig.~\ref{fig:qk}, we can see that applying output-invariant weight scaling causes PCS-based and ESD scale-based similarity to degrade substantially as we use larger scaling factors, while the spectral signature with \ALPHAHILL remains stable, correctly recognizing the transformed weights as functionally identical. Furthermore, under MLP hidden-unit permutations, PCS collapses to near zero, but \ALPHAHILL remains near-perfect (Tab.~\ref{tab:MLP}).

Here, we discuss why this ablation is important.
Data-aware similarity (e.g., Logits and REEF) is expected to remain high under strictly output-invariant transforms, but it requires a prompt distribution and $O(\#\text{prompts})$ inference per comparison, making it unsuitable as a zoo-scale kernel. Weight-matrix fingerprints such as \textsc{AWM}~\cite{zeng2025awm} reduce some reparameterization sensitivity but still require matrix-level comparisons (and often assignment/matching) per pair. In contrast, our ESD shape signature is computed once per model and yields invariance to common output-preserving symmetries without explicit matching. Note that pure scale metrics (Log Norm, Spectral Norm) are also invariant here, but precisely because they track magnitude, they tend to saturate and are less discriminative across independently trained families (see Sec.~\ref{sec:result_llm_lineage}).

\begin{figure}[ht!]
    \centering
    \includegraphics[width=\linewidth]{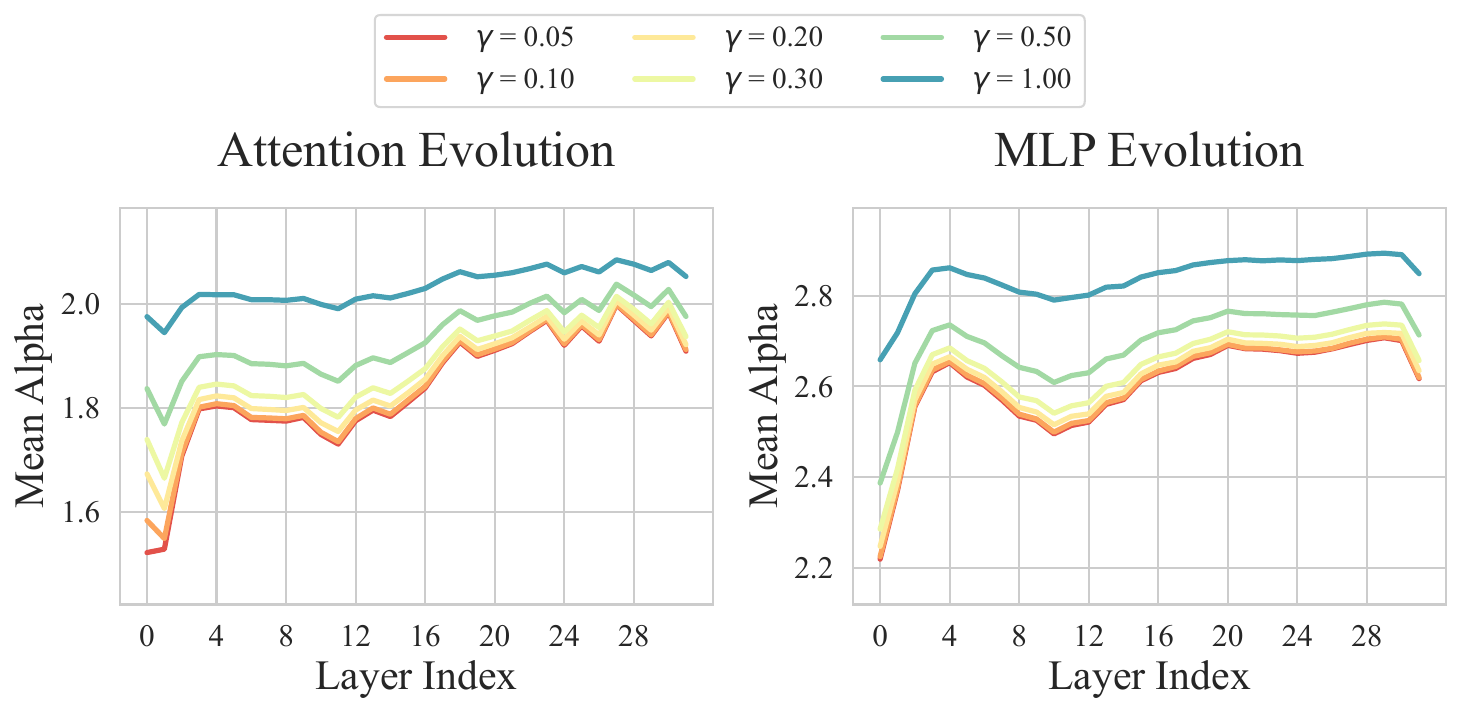}
    \caption{
    Layer-wise evolution of \ALPHAHILL under random perturbations. Curves show the \ALPHAHILL across attention and MLP modules for different noise levels $\gamma$.
    }
    \Description{}
    \label{fig:noise_evol}
\end{figure}

\textbf{\ALPHAHILL remains robust under random perturbations.}
To simulate post-training weight changes, we inject Gaussian noise of increasing magnitude into model weight entries and evaluate the stability of different similarity metrics. We follow the noise strength $\gamma$ definition in Appendix~\ref{ablation_app} and report full results in Tab.~\ref{tab:ablation_noise}. Across all noise levels, our \ALPHAHILL–based similarity remains close to the unperturbed model, but PCS degrades substantially as noise magnitude increases. These results indicate that spectral shape features provide a robust measure of model similarity under realistic weight-space perturbations. Moreover, spectral scale metrics also perform well in this scenario since additive Gaussian noise perturbs weights without strongly changing their overall magnitude. Beyond global averages, we also analyze the layer-wise response to perturbation. Fig.~\ref{fig:noise_evol} demonstrates that larger perturbations generally increase \ALPHAHILL while preserving similar depth-wise trends, showing that spectral signatures respond systematically to weight-space noise.

\begin{figure}
    \centering
    \includegraphics[width=0.8\linewidth]{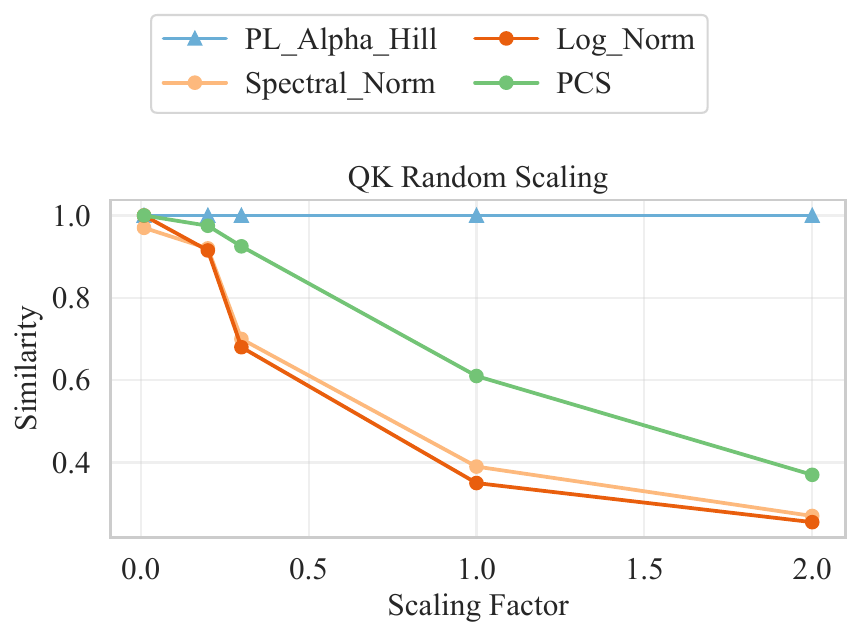}
    \caption{Robustness of different similarity metrics under output-invariant scaling to consecutive Query and Key modules. \ALPHAHILL is a shape metric, while Log Norm and Spectral Norm are scale metrics.}    \Description{}
    \label{fig:qk}
\end{figure}

\begin{table}[!ht]
\centering
\caption{Robustness of different metrics under hidden-unit permutation.}
\label{tab:MLP}
\resizebox{0.65\columnwidth}{!}{%
    \begin{tabular}{lcc}
    \toprule
    \textbf{Metric} & \textbf{MLP-In} & \textbf{MLP-Out} \\
    \midrule
    PCS & 0.00 & 0.00 \\
    Log\_Norm             & 1.00 & 1.00 \\
    Spectral\_Norm        & 1.00 & 1.00 \\
    Log\_Alpha\_Norm       & 1.00 & 1.00 \\
    \rowcolor{black!10} \ALPHAHILL (ours)     & 1.00 & 1.00 \\
    \bottomrule
    \end{tabular}%
}
\end{table}

\begin{table}[!ht]
\centering
\caption{
Similarities of distinct metrics under increasing noise.
}
\label{tab:ablation_noise}
\resizebox{1.0\columnwidth}{!}{%
    \begin{tabular}{lccccc}
    \toprule
    \textbf{Metric} & $\boldsymbol{\gamma}~\mathbf{=0.1}$ & $\boldsymbol{\gamma}~\mathbf{=0.3}$ & $\boldsymbol{\gamma}~\mathbf{=0.5}$ & $\boldsymbol{\gamma}~\mathbf{=1.0}$ & $\boldsymbol{\gamma}~\mathbf{=2.0}$ \\
    \midrule
    PCS                 & 1.00 & 0.96 & 0.89 & 0.70 & 0.44 \\
    Log\_Norm           & 1.00 & 1.00 & 1.00 & 1.00 & 1.00 \\
    Spectral\_Norm      & 1.00 & 1.00 & 1.00 & 0.98 & 0.89 \\
    Log\_Alpha\_Norm    & 0.99 & 0.87 & 0.85 & 0.81 & 0.81 \\
    \rowcolor{black!10} \ALPHAHILL (ours)  & 1.00 & 0.99 & 0.98 & 0.98 & 0.96 \\
    \bottomrule
    \end{tabular}
}
\end{table}

\section{Conclusion}
\label{s:conclusion}

This work presents a spectral shape–based framework for large-scale analysis of LLMs that is data-free, computationally efficient, and robust to post-training transformations. We curate and benchmark a large and diverse LLM corpus and conduct extensive experiments across similarity measurement, classification, clustering, and performance prediction. Our results have shown the consistent effectiveness of our spectral signature across various experimental setups. Furthermore, we validate that ESD shape metrics, particularly \ALPHAHILL, capture intrinsic structural properties of LLMs that are stable under output-invariant reparameterizations and noise perturbations. Therefore, our spectral signature can provide a practical tool for managing modern LLM ecosystems, with implications for model provenance, organization, and scalable evaluation.

\begin{acks}
YY would like to acknowledge the support of the U.S. Department of Energy (Grant number: DE-SC0025584) and the Defense Advanced Research Projects Agency (Grant number: HR00112520011). HL was supported by the U.S. Department of Energy under Contract DE-AC02-05CH11231
and U.S. National Science Foundation NSF-DMS 2412403. We also would like to acknowledge NERSC DOE Mission Science Allocation for ERCAP request ERCAP0035003.
\end{acks}

\clearpage
\bibliographystyle{ACM-Reference-Format}
\balance
\bibliography{refs}

@article{zhou2023temperature,
  title={Temperature balancing, layer-wise weight analysis, and neural network training},
  author={Zhou, Yefan and Pang, Tianyu and Liu, Keqin and Mahoney, Michael W and Yang, Yaoqing and others},
  journal={Advances in Neural Information Processing Systems},
  volume={36},
  pages={63542--63572},
  year={2023}
}

@misc{sun2025idiosyncrasies,
  title={Idiosyncrasies in large language models},
  author={Sun, Mingjie and Yin, Yida and Xu, Zhiqiu and Kolter, J Zico and Liu, Zhuang},
  journal={arXiv preprint arXiv:2502.12150},
  year={2025}
}

@inproceedings{zeng2024HuRef,
  title     = {{HuRef}: {HU}man-{RE}adable Fingerprint for Large Language Models},
  author    = {Zeng, Boyi and Wang, Lizheng and Hu, Yuncong and Xu, Yi and Zhou, Chenghu and Wang, Xinbing and Yu, Yu and Lin, Zhouhan},
  booktitle = {Advances in Neural Information Processing Systems},
  volume    = {37},
  pages     = {126332--126362},
  year      = {2024},
  publisher = {Curran Associates, Inc.},
  address ={Vancouver, Canada}
}

@misc{horwitz2025Atlas,
    author = {Horwitz, Eliahu and Kurer, Nitzan and Kahana, Jonathan and Amar, Liel and Hoshen, Yedid},
    title = {We Should Chart an Atlas of All the World’s Models},
    journal = {arXiv preprint arXiv:2503.10633v2},
    year = {2025}
}

@inproceedings{horwitz2025TreeHeritage,
  author    = {Eliahu Horwitz and
               Asaf Shul and
               Yedid Hoshen},
  title     = {Unsupervised Model Tree Heritage Recovery},
  booktitle = {Proceedings of the 13th International Conference on Learning Representations},
  publisher = {OpenReview.net},
  year      = {2025},
  pages={47900--47918},
  address   = {Singapore}
}

@misc{gueta2023KnowledgeRegion,
    author = {Gueta, Almog and Venezian, Elad and Raffel, Colin and Slonim, Noam and Katz, Yoav and Choshen, Leshem},
    title = {Knowledge is a Region in Weight Space for Finetuned Language Models},
    journal = {arXiv preprint arXiv:2302.04863v3},
    year = {2023}
}

@misc{schurholt2025ModelZoo,
    author = {Schürholt, Konstantin and Meynent, Léo and Zhou, Yefan and Lu, Haiquan and Yang, Yaoqing and Borth, Damian},
    title = {A Model Zoo on Phase Transitions in Neural Networks},
    journal = {arXiv preprint arXiv:2504.18072v1},
    year = {2025}
}

@inproceedings{Yang2021TaxonomizingLossLandscapes,
    author = {Yang, Yaoqing and Hodgkinson, Liam and Theisen, Ryan and Zou, Joe and Gonzalez, Joseph E and Ramchandran, Kannan and Mahoney, Michael W},
    title = {Taxonomizing local versus global structure in
    neural network loss landscapes},
    booktitle = {Advances in Neural Information Processing Systems},
    volume = {34},
    publisher = {Curran Associates, Inc.},
    pages = {18722--18733},
    year = {2021},
    address={Virtual}
}

@inproceedings{mahoney2019traditional,
  title={Traditional and heavy tailed self regularization in neural network models},
  author={Mahoney, Michael and Martin, Charles},
  booktitle={International Conference on Machine Learning},
  pages={4284--4293},
  year={2019},
  publisher={PMLR},
  address={Long Beach, CA, USA}
}

@article{martin2021implicit,
  title={Implicit self-regularization in deep neural networks: Evidence from random matrix theory and implications for learning},
  author={Martin, Charles H and Mahoney, Michael W},
  journal={Journal of Machine Learning Research},
  volume={22},
  number={165},
  pages={1--73},
  year={2021}
}

@book{couillet2022random,
  title={Random matrix methods for machine learning},
  author={Couillet, Romain and Liao, Zhenyu},
  year={2022},
  publisher={Cambridge University Press},
  address = {Cambridge, UK}
}

@misc{liao2025random,
  title={Random Matrix Theory for Deep Learning: Beyond Eigenvalues of Linear Models},
  author={Liao, Zhenyu and Mahoney, Michael W},
  journal={arXiv preprint arXiv:2506.13139},
  year={2025}
}

@inproceedings{liu-etal-2024-model,
    title = "Model Balancing Helps Low-data Training and Fine-tuning",
    author = "Liu, Zihang  and
      Hu, Yuanzhe  and
      Pang, Tianyu  and
      Zhou, Yefan  and
      Ren, Pu  and
      Yang, Yaoqing",
    editor = "Al-Onaizan, Yaser  and
      Bansal, Mohit  and
      Chen, Yun-Nung",
    booktitle = "Proceedings of the 2024 Conference on Empirical Methods in Natural Language Processing",
    month = nov,
    year = "2024",
    address = "Miami, Florida, USA",
    publisher = "Association for Computational Linguistics",
    url = "https://aclanthology.org/2024.emnlp-main.78/",
    doi = "10.18653/v1/2024.emnlp-main.78",
    pages = "1311--1331",
}

@inproceedings{hu2025eigenspectrum,
  author    = {Yuanzhe Hu and
               Kinshuk Goel and
               Vlad Killiakov and
               Yaoqing Yang},
  title     = {Eigenspectrum Analysis of Neural Networks without Aspect Ratio Bias},
  booktitle = {Proceedings of the 42nd International Conference on Machine Learning},
  series    = {Proceedings of Machine Learning Research},
  publisher = {PMLR},
  year      = {2025},
  pages={24290-24313},
  address   = {Vancouver, Canada}
}

@article{lu2024alphapruning,
  title={Alphapruning: Using heavy-tailed self regularization theory for improved layer-wise pruning of large language models},
  author={Lu, Haiquan and Zhou, Yefan and Liu, Shiwei and Wang, Zhangyang and Mahoney, Michael W and Yang, Yaoqing},
  journal={Advances in neural information processing systems},
  volume={37},
  pages={9117--9152},
  year={2024}
}

@inproceedings{gurbuzbalaban2021heavy,
  title={The heavy-tail phenomenon in SGD},
  author={Gurbuzbalaban, Mert and Simsekli, Umut and Zhu, Lingjiong},
  booktitle={International Conference on Machine Learning},
  pages={3964--3975},
  year={2021},
  publisher={PMLR},
  address={Virtual}
}

@inproceedings{hodgkinson2021multiplicative,
  title={Multiplicative noise and heavy tails in stochastic optimization},
  author={Hodgkinson, Liam and Mahoney, Michael},
  booktitle={International Conference on Machine Learning},
  pages={4262--4274},
  year={2021},
  address={Virtual},
  publisher={PMLR}
}

@inproceedings{hodgkinson2022generalization,
  title={Generalization bounds using lower tail exponents in stochastic optimizers},
  author={Hodgkinson, Liam and Simsekli, Umut and Khanna, Rajiv and Mahoney, Michael},
  booktitle={International Conference on Machine Learning},
  pages={8774--8795},
  year={2022},
  address={Baltimore, Maryland, USA},
  publisher={PMLR}
}

@inproceedings{simsekli2019tail,
  title={A tail-index analysis of stochastic gradient noise in deep neural networks},
  author={Simsekli, Umut and Sagun, Levent and Gurbuzbalaban, Mert},
  booktitle={International Conference on Machine Learning},
  pages={5827--5837},
  year={2019},
  publisher={PMLR},
  address={Long Beach, CA, USA}
}

@article{simsekli2020hausdorff,
  title={Hausdorff dimension, heavy tails, and generalization in neural networks},
  author={Simsekli, Umut and Sener, Ozan and Deligiannidis, George and Erdogdu, Murat A},
  journal={Advances in Neural Information Processing Systems},
  volume={33},
  pages={5138--5151},
  year={2020}
}

@inproceedings{yang2023test,
  title={Test accuracy vs. generalization gap: Model selection in nlp without accessing training or testing data},
  author={Yang, Yaoqing and Theisen, Ryan and Hodgkinson, Liam and Gonzalez, Joseph E and Ramchandran, Kannan and Martin, Charles H and Mahoney, Michael W},
  booktitle={Proceedings of the 29th ACM SIGKDD Conference on Knowledge Discovery and Data Mining},
  pages={3011--3021},
  year={2023},
  publisher = {ACM},
  address   = {Long Beach, CA, USA}
}

@article{hill1975simple,
  title={A simple general approach to inference about the tail of a distribution},
  author={Hill, Bruce M},
  journal={The annals of statistics},
  volume={3},
  number={5},
  pages={1163--1174},
  year={1975},
  publisher={JSTOR}
}

@inproceedings{jin2024proflingo,
  title={Proflingo: A fingerprinting-based intellectual property protection scheme for large language models},
  author={Jin, Heng and Zhang, Chaoyu and Shi, Shanghao and Lou, Wenjing and Hou, Y Thomas},
  booktitle={2024 IEEE Conference on Communications and Network Security (CNS)},
  pages={1--9},
  year={2024},
  publisher={IEEE},
  address   = {Taipei, Taiwan}
}

@misc{zhuang2025embedllm,
      title={EmbedLLM: Learning Compact Representations of Large Language Models}, 
      author={Richard Zhuang and Tianhao Wu and Zhaojin Wen and Andrew Li and Jiantao Jiao and Kannan Ramchandran},
      year={2024},
      eprint={2410.02223},
      archivePrefix={arXiv},
      primaryClass={cs.CL},
      url={https://arxiv.org/abs/2410.02223}, 
}

@inproceedings{zhu2025independence,
  author       = {Sally Zhu and
                  Ahmed M. Ahmed and
                  Rohith Kuditipudi and
                  Percy Liang},
  title        = {Independence Tests for Language Models},
  booktitle = {Proceedings of the 42nd International Conference on Machine Learning},
  series    = {Proceedings of Machine Learning Research},
  publisher = {PMLR},
  year      = {2025},
  pages={79673-79698},
  address   = {Vancouver, Canada}
}

@inproceedings{Nikolic2025model,
    title={Model Provenance Testing for Large Language Models}, 
    author={Ivica Nikolic and Teodora Baluta and Prateek Saxena},
    booktitle = {Advances in Neural Information Processing Systems},
    publisher = {Curran Associates, Inc.},    
    pages = {34126--34153},
    address={Sydney, Australia},
    year = {2025}
}

@inproceedings{wu-etal-2023-llmdet,
     title = "{LLMD}et: A Third Party Large Language Models Generated Text Detection Tool",
    author = "Wu, Kangxi  and
      Pang, Liang  and
      Shen, Huawei  and
      Cheng, Xueqi  and
      Chua, Tat-Seng",
    editor = "Bouamor, Houda  and
      Pino, Juan  and
      Bali, Kalika",
    booktitle = "Findings of the Association for Computational Linguistics: EMNLP 2023",
    month = dec,
    year = "2023",
    address = "Singapore",
    publisher = "Association for Computational Linguistics",
    url = "https://aclanthology.org/2023.findings-emnlp.139/",
    doi = "10.18653/v1/2023.findings-emnlp.139",
    pages = "2113--2133"
}

@inproceedings{
jiang2025stealthink,
title={StealthInk: A Multi-bit and Stealthy Watermark for Large Language Models},
author={Ya Jiang and Chuxiong Wu and Massieh Kordi Boroujeny and Brian Mark and Kai Zeng},
  booktitle = {Proceedings of the 42nd International Conference on Machine Learning},
  series    = {Proceedings of Machine Learning Research},
  publisher = {PMLR},
  year      = {2025},
  pages={27685-27709},
  address   = {Vancouver, Canada}
}

@misc{Gu2022WatermarkingPL,
  title={Watermarking Pre-trained Language Models with Backdooring},
  author={Chenxi Gu and Chengsong Huang and Xiaoqing Zheng and Kai-Wei Chang and Cho-Jui Hsieh},
  journal={ArXiv},
  year={2022},
  volume={abs/2210.07543},
  url={https://api.semanticscholar.org/CorpusID:252907247}
}

@inproceedings{wang2019attacks,
  author    = {Wang, Tianhao and Kerschbaum, Florian},
  title     = {Attacks on Digital Watermarks for Deep Neural Networks},
  booktitle = {2019 IEEE International Conference on Acoustics, Speech and Signal Processing (ICASSP)},
  pages     = {2622--2626},
  year      = {2019},
  publisher = {IEEE},
  address   = {Brighton, UK},
  doi       = {10.1109/ICASSP.2019.8682202}
}

@inproceedings{Uchida2017embedding,
author = {Uchida, Yusuke and Nagai, Yuki and Sakazawa, Shigeyuki and Satoh, Shin'ichi},
title = {Embedding Watermarks into Deep Neural Networks},
year = {2017},
isbn = {9781450347013},
publisher = {Association for Computing Machinery},
address = {New York, NY, USA},
url = {https://doi.org/10.1145/3078971.3078974},
doi = {10.1145/3078971.3078974},
booktitle = {Proceedings of the 2017 ACM on International Conference on Multimedia Retrieval},
pages = {269–277},
numpages = {9},
location = {Bucharest, Romania},
series = {ICMR '17}
}

@article{martin2020predicting_NatComm,
  title={Predicting trends in the quality of state-of-the-art neural networks without access to training or testing data},
  author={Martin, Charles H and Peng, Tongsu Serena and Mahoney, Michael W},
  journal={Nature Communications},
  volume={12},
  number={1},
  pages={1--13},
  year={2021},
  publisher={Nature Publishing Group}
}

@misc{martin2025setol,
  title={SETOL: A Semi-Empirical Theory of (Deep) Learning},
  author={Martin, Charles H and Hinrichs, Christopher},
  journal={arXiv preprint arXiv:2507.17912},
  year={2025}
}

@misc{yoon2025intrinsic,
  title={Intrinsic Fingerprint of LLMs: Continue Training is NOT All You Need to Steal A Model!},
  author={Yoon, Do-hyeon and Chun, Minsoo and Allen, Thomas and M{\"u}ller, Hans and Wang, Min and Sharma, Rajesh},
  journal={arXiv preprint arXiv:2507.03014},
  year={2025}
}

@misc{wu2025llm,
  title={LLM DNA: Tracing Model Evolution via Functional Representations},
  author={Wu, Zhaomin and Zhao, Haodong and Wang, Ziyang and Guo, Jizhou and Wang, Qian and He, Bingsheng},
  journal={arXiv preprint arXiv:2509.24496},
  year={2025}
}

@misc{wang2025ghost,
  title={Ghost in the Transformer: Tracing LLM Lineage with SVD-Fingerprint},
  author={Wang, Suqing and Ma, Ziyang and Li, Xinyi and Li, Zuchao},
  journal={arXiv preprint arXiv:2511.06390},
  year={2025}
}

@misc{zeng2025awm,
  title={AWM: Accurate Weight-Matrix Fingerprint for Large Language Models},
  author={Zeng, Boyi and Chen, Lin and He, Ziwei and Wang, Xinbing and Lin, Zhouhan},
  journal={arXiv preprint arXiv:2510.06738},
  year={2025}
}

@misc{myrzakhan2024openllmleaderboard,
      title={Open-LLM-Leaderboard: From Multi-choice to Open-style Questions for LLMs Evaluation, Benchmark, and Arena}, 
      author={Myrzakhan, Aidar and Mahmoud Bsharat, Sondos and Shen, Zhiqiang},
      year={2024},
      journal={arXiv preprint arXiv:2406.07545}
}

@misc{hendrycks2021measuring,
      title={Measuring Massive Multitask Language Understanding}, 
      author={Dan Hendrycks and Collin Burns and Steven Basart and Andy Zou and Mantas Mazeika and Dawn Song and Jacob Steinhardt},
      year={2021},
      eprint={2009.03300},
      archivePrefix={arXiv},
      primaryClass={cs.CY},
      url={https://arxiv.org/abs/2009.03300},
}

@inproceedings{zellers-etal-2019-hellaswag,
    title = "{H}ella{S}wag: Can a Machine Really Finish Your Sentence?",
    author = "Zellers, Rowan  and
      Holtzman, Ari  and
      Bisk, Yonatan  and
      Farhadi, Ali  and
      Choi, Yejin",
    editor = "Korhonen, Anna  and
      Traum, David  and
      M{\`a}rquez, Llu{\'i}s",
    booktitle = "Proceedings of the 57th Annual Meeting of the Association for Computational Linguistics",
    month = jul,
    year = "2019",
    address = "Florence, Italy",
    publisher = "Association for Computational Linguistics",
    url = "https://aclanthology.org/P19-1472/",
    doi = "10.18653/v1/P19-1472",
    pages = "4791--4800"
}

@misc{clark2018thinksolvedquestionanswering,
      title={Think you have Solved Question Answering? Try ARC, the AI2 Reasoning Challenge}, 
      author={Peter Clark and Isaac Cowhey and Oren Etzioni and Tushar Khot and Ashish Sabharwal and Carissa Schoenick and Oyvind Tafjord},
      year={2018},
      eprint={1803.05457},
      journal={arXiv},
      primaryClass={cs.AI},
      url={https://arxiv.org/abs/1803.05457}, 
}

@inproceedings{zhang2025reef,
  author       = {Jie Zhang and
                  Dongrui Liu and
                  Chen Qian and
                  Linfeng Zhang and
                  Yong Liu and
                  Yu Qiao and
                  Jing Shao},
  title        = {{REEF:} Representation Encoding Fingerprints for Large Language Models},
  booktitle    = {International Conference on Learning Representations},
  publisher    = {OpenReview.net},
  year         = {2025},
  pages={48092-48117},
  address={Singapore}
}

@book{kendall1948rank,
  title     = {Rank Correlation Methods},
  author    = {Kendall, Maurice George},
  year      = {1948},
  publisher = {Charles Griffin \& Company},
  address   = {London, UK}
}

@article{rousseeuw1987silhouettes,
  title={Silhouettes: a graphical aid to the interpretation and validation of cluster analysis},
  author={Rousseeuw, Peter J},
  journal={Journal of computational and applied mathematics},
  volume={20},
  pages={53--65},
  year={1987},
  publisher={Elsevier}
}

@article{davies2009cluster,
  title={A cluster separation measure},
  author={Davies, David L and Bouldin, Donald W},
  journal={IEEE transactions on pattern analysis and machine intelligence},
  volume={PAMI-1},
  number={2},
  pages={224--227},
  year={2009},
  publisher={Ieee}
}

@inproceedings{lin2022truthfulqa,
  title={Truthfulqa: Measuring how models mimic human falsehoods},
  author={Lin, Stephanie and Hilton, Jacob and Evans, Owain},
  booktitle={Proceedings of the 60th annual meeting of the association for computational linguistics (volume 1: long papers)},
  pages={3214--3252},
  year={2022},
  publisher = {Association for Computational Linguistics},
  address={Dublin, Ireland}
}

@misc{Yang2024AFF,
  title={A Fingerprint for Large Language Models},
  author={Zhiguang Yang and Hanzhou Wu},
  journal={ArXiv},
  year={2024},
  volume={abs/2407.01235},
  url={https://api.semanticscholar.org/CorpusID:270869924}
}

@software{openlm2023openllama,
  author = {Geng, Xinyang and Liu, Hao},
  title = {OpenLLaMA: An Open Reproduction of LLaMA},
  month = May,
  year = 2023,
  organization={OpenLM Research},
  url = {https://github.com/openlm-research/open_llama}
}

@misc{bai2023qwentechnicalreport,
      title={Qwen Technical Report}, 
      author={Jinze Bai and Shuai Bai and Yunfei Chu and others},
      year={2023},
      eprint={2309.16609},
      archivePrefix={arXiv},
      primaryClass={cs.CL},
      url={https://arxiv.org/abs/2309.16609}, 
}

@misc{jiang2023mistral7b,
      title={Mistral 7B}, 
      author={Albert Q. Jiang and Alexandre Sablayrolles and Arthur Mensch and others},
      year={2023},
      eprint={2310.06825},
      archivePrefix={arXiv},
      primaryClass={cs.CL},
      url={https://arxiv.org/abs/2310.06825}, 
}

@misc{touvron2023llamaopenefficientfoundation,
      title={LLaMA: Open and Efficient Foundation Language Models}, 
      author={Hugo Touvron and Thibaut Lavril and Gautier Izacard and others},
      year={2023},
      eprint={2302.13971},
      archivePrefix={arXiv},
      primaryClass={cs.CL},
      url={https://arxiv.org/abs/2302.13971}, 
}

@misc{touvron2023llama2openfoundation,
      title={Llama 2: Open Foundation and Fine-Tuned Chat Models}, 
      author={Hugo Touvron and Louis Martin and Kevin Stone and others},
      year={2023},
      eprint={2307.09288},
      archivePrefix={arXiv},
      primaryClass={cs.CL},
      url={https://arxiv.org/abs/2307.09288}, 
}

@misc{grattafiori2024llama3herdmodels,
      title={The Llama 3 Herd of Models}, 
      author={Aaron Grattafiori and Abhimanyu Dubey and Abhinav Jauhri and others},
      year={2024},
      eprint={2407.21783},
      archivePrefix={arXiv},
      primaryClass={cs.AI},
      url={https://arxiv.org/abs/2407.21783}, 
}

@misc{tang2025pangupromoemixture,
      title={Pangu Pro MoE: Mixture of Grouped Experts for Efficient Sparsity}, 
      author={Yehui Tang and Xiaosong Li and Fangcheng Liu and Wei Guo and Hang Zhou and Yaoyuan Wang and Kai Han and Xianzhi Yu and Jinpeng Li and Hui Zang and Fei Mi and Xiaojun Meng and Zhicheng Liu and Hanting Chen and Binfan Zheng and Can Chen and Youliang Yan and Ruiming Tang and Peifeng Qin and Xinghao Chen and Dacheng Tao and Yunhe Wang},
      year={2025},
      eprint={2505.21411},
      archivePrefix={arXiv},
      primaryClass={cs.CL},
      url={https://arxiv.org/abs/2505.21411}, 
}

@article{DP,
author = {Richard Bellman},
title = {Dynamic Programming},
journal = {Science},
volume = {127},
number = {3304},
pages = {976-976},
year = {1958},
}

@article{alstott2014powerlaw,
  title={powerlaw: a Python package for analysis of heavy-tailed distributions},
  author={Alstott, Jeff and Bullmore, Ed and Plenz, Dietmar},
  journal={PloS one},
  volume={9},
  number={1},
  pages={e85777},
  year={2014},
  publisher={Public Library of Science San Francisco, USA}
}

@article{clauset2009power,
  title={Power-law distributions in empirical data},
  author={Clauset, Aaron and Shalizi, Cosma Rohilla and Newman, Mark EJ},
  journal={SIAM review},
  volume={51},
  number={4},
  pages={661--703},
  year={2009},
  publisher={SIAM}
}

@misc{hodgkinson2025models,
  title={Models of heavy-tailed mechanistic universality},
  author={Hodgkinson, Liam and Wang, Zhichao and Mahoney, Michael W},
  journal={arXiv preprint arXiv:2506.03470},
  year={2025}
}

@article{marvcenko1967distribution,
  title={Distribution of eigenvalues for some sets of random matrices},
  author={Mar{\v{c}}enko, Vladimir A and Pastur, Leonid Andreevich},
  journal={Mathematics of the USSR-Sbornik},
  volume={1},
  number={4},
  pages={457--483},
  year={1967}
}

\section{Appendix}
\label{sec:appendix}

\subsection{Robustness of ESD Shape Metrics}
\label{app:robust_esd}

Pre-training and post-training reshape the correlations within weight matrices, as reflected in variations in the heavy-tailed index across layers. To validate the reliability of our method, we investigate whether the spectral structure preserves sufficient lineage information under post-training perturbations.

\begin{figure}
    \centering
    \includegraphics[width=1.0\linewidth]{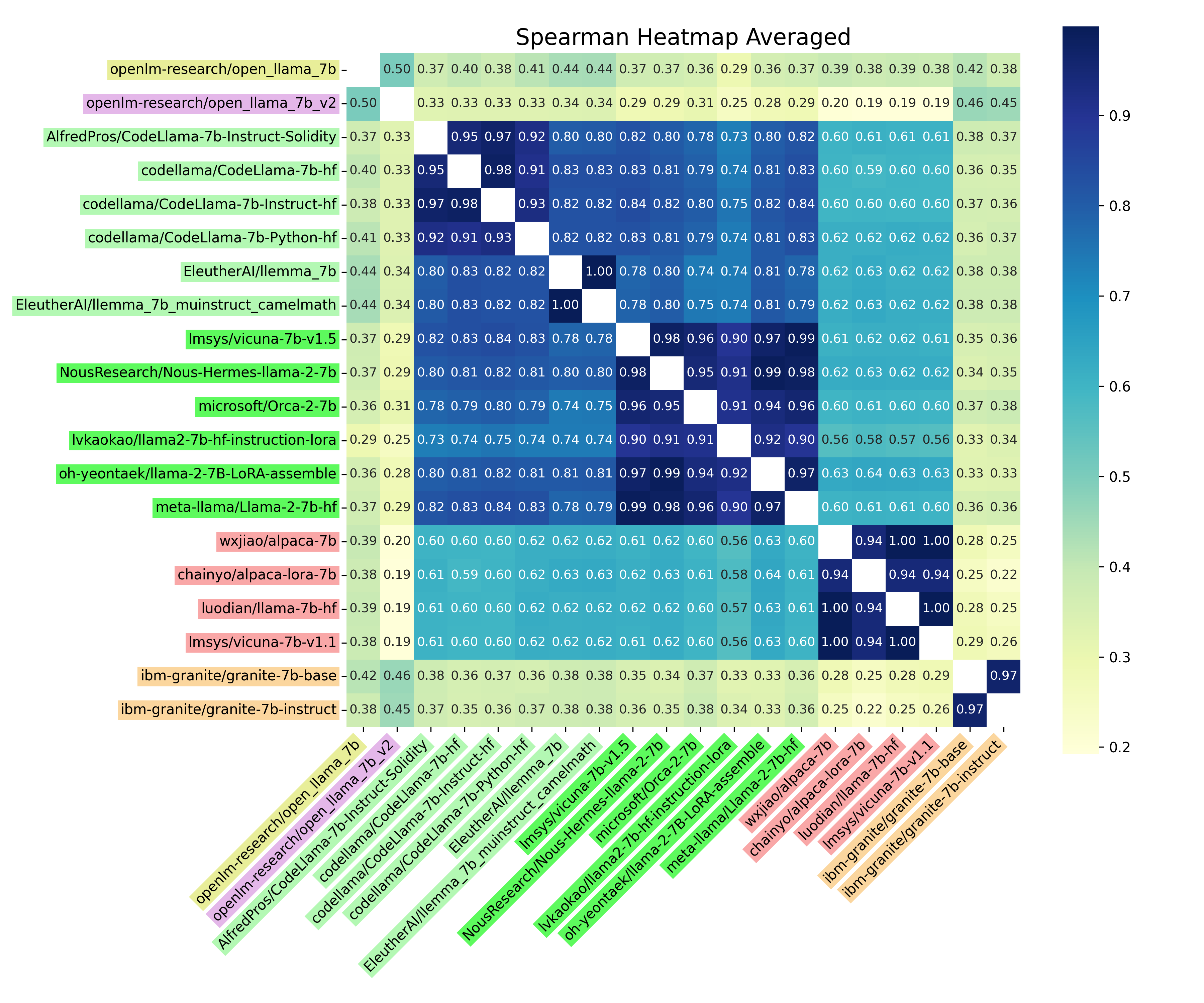}
    \caption{
    Heatmap of spectral signature similarity across models derived from distinct 7B base models, measured by Spearman's rank correlation of \ALPHAHILL. Axis colors indicate base-model lineage; dark and light green denote Llama-2-7B and CodeLlama-7B-hf derivatives, respectively. The clear block structure shows that spectral signatures preserve ancestry despite diverse fine-tuning trajectories.}
    \label{fig:heatmap}
\end{figure}

\paragraph{\ALPHAHILL is robust under perturbation.} Fig.~\ref{fig:heatmap} shows that ESD shape metrics can encode lineage information beyond pretraining.
Using Spearman correlation between layer-wise \ALPHAHILL profiles, models derived from the same 7B base model exhibit high similarities, whereas similarities across base families are substantially lower.
Moreover, within the Llama-2 lineage, we observe finer sub-structure: distinct clusters align with different post-training trajectories (e.g., instruction-tuned vs. CodeLlama-style derivatives), suggesting that post-training alters spectral signatures in a structured manner. These observations motivate our use of spectral signatures for downstream tasks such as lineage attribution, clustering, and performance prediction (Sec.\ref{sec:results}).

\begin{figure}
   \centering
   \includegraphics[width=0.98\linewidth]{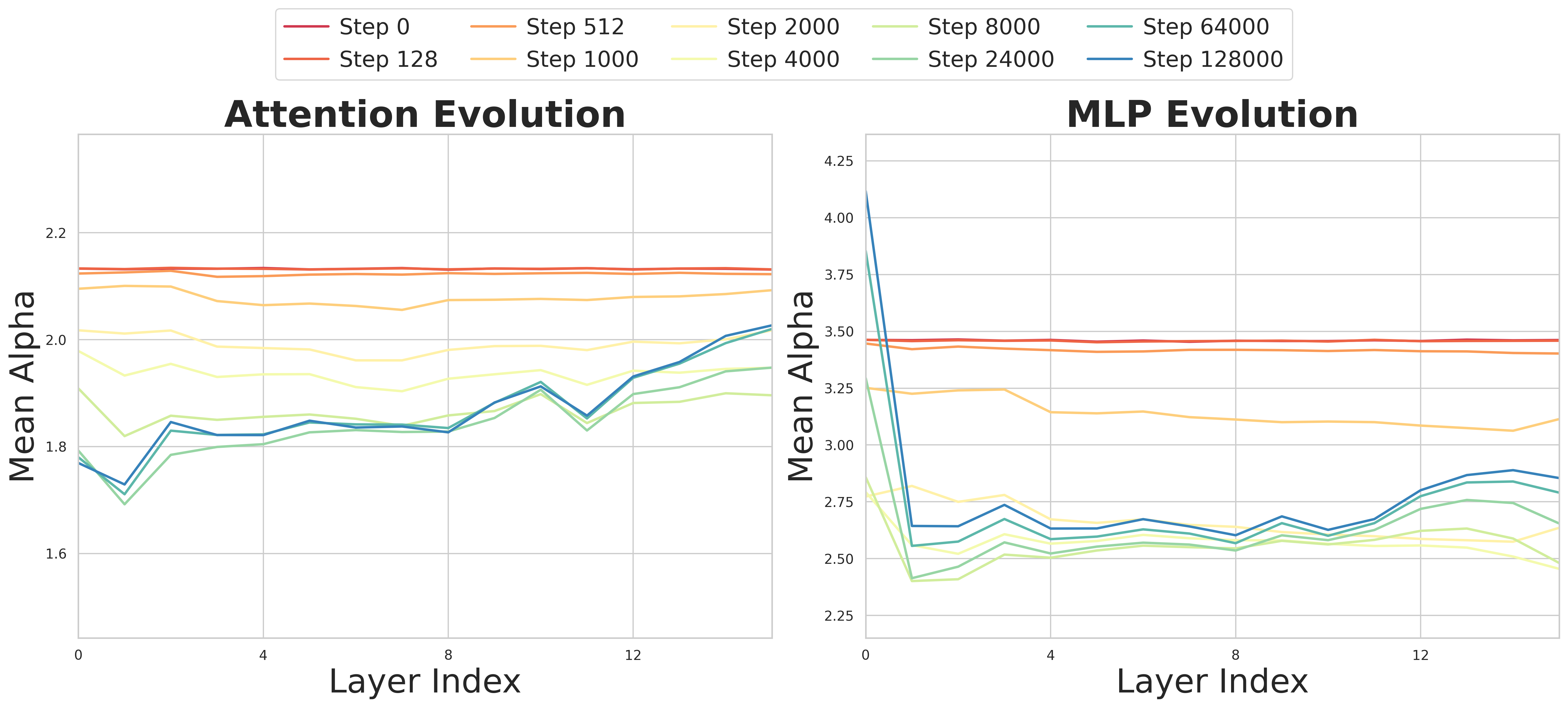}
   \caption{The evolution of layer-wise \ALPHAHILL through the training process. The values of \ALPHAHILL are averaged within Attention modules and MLP modules.}
   \label{fig:forge}
\end{figure}

\paragraph{The Forging of Spectral Signature}
As illustrated in Fig. ~\ref{fig:forge}, we observe three distinct phases of spectral evolution: 
\begin{enumerate}
\item \textbf{Initialization.} At Step 0, all layers exhibit large, uniform \ALPHAHILL values (metrics for attention and MLP weight matrices differ due to the aspect ratio difference~\cite{hu2025eigenspectrum}), resembling MP distribution for random matrices~\citep{marvcenko1967distribution}.
\item \textbf{Formulation.} As training progresses, \ALPHAHILL decreases globally, signifying feature learning. Crucially, \ALPHAHILL drops in early layers, flattens in middle layers, dips in mid-deep layers, and rises in the final layers.
\item \textbf{Stabilization} As shown in Fig. ~\ref{fig:forge}, the spectral signature stabilizes rapidly. While the early phase ($<16k$ steps) shows high volatility, the signature ``locks in'' after step 16,000. Subsequent updates refine the weights but do not alter the fundamental topological shape. 
\end{enumerate}

\noindent\paragraph{\ALPHAHILL undergo minimal change during post-training.}
Because post-training updates (e.g., SFT, RLVR) introduce relatively small changes in parameter space, they do not erase the heavy-tailed structure established during pre-training. Based on that, \ALPHAHILL\ works as a stable and lineage-preserving signature that remains robust to common downstream adaptations. This motivates us to use it for performance prediction and model attribution.

\subsection{Metrics for Classification Quality}\label{app:classification_metric}
Given two signatures $\mathbf{Z}_1,\mathbf{Z}_2$, we consider:
(i) Spearman rank correlation $\rho_s(\mathbf{Z}_1,\mathbf{Z}_2)$ as our default similarity (Eq.~\ref{eq:sr}),
(ii) an RBF kernel applied to the signature distance, 
\begin{equation}
k_{\text{RBF}}(\mathbf{Z}_1,\mathbf{Z}_2)=\exp\!\left(-\frac{\|\mathbf{Z}_1-\mathbf{Z}_2\|_2^2}{2\sigma^2}\right)
\label{eq:rbf}
\end{equation}
, and
(iii) cosine similarity,
\begin{equation}
k_{\cos}\mathbf({Z}_1,\mathbf{Z}_2)=\frac{\langle \mathbf{Z}_1,\mathbf{Z}_2\rangle}{\|\mathbf{Z}_1\|_2\|\mathbf{Z}_2\|_2}.
\label{eq:cos}
\end{equation}

\subsection{Metrics for Internal Clustering Validation}
\label{app:eval_metrics_clustering}

For unsupervised model-family discovery, we evaluate spectral-signature clusters using two standard internal validity indices: the Silhouette Coefficient and the Davies-Bouldin Index.

\paragraph{Silhouette Coefficient.}
The Silhouette Coefficient measures clustering quality by comparing intra-cluster cohesion and inter-cluster separation. For a point $i \in C_k$, let $a(i)$ denote its average distance to other points in $C_k$, and let $b(i)$ denote the smallest average distance to points in any other cluster. The silhouette coefficient is
\begin{equation}
s(i) = \frac{b(i) - a(i)}{\max{a(i), b(i)}} .
\end{equation}
The overall Silhouette Score is the average of $s(i)$ over all samples.

\paragraph{Davies-Bouldin Index.}
The Davies-Bouldin Index (DBI) compares intra-cluster dispersion with inter-cluster separation. Let $S_i$ denote the dispersion of cluster $i$, and $M_{ij}$ denote the distance between the centroids of clusters $i$ and $j$. Define
\begin{equation}
R_{ij} = \frac{S_i + S_j}{M_{ij}}.
\end{equation}
Then
\begin{equation}
DB = \frac{1}{k} \sum_{i=1}^{k} \max_{j \neq i} R_{ij}.
\end{equation}

\subsection{Additional Results on Model Clustering}\label{app:cluster}

\begin{figure}[!ht]    
    \centering
    \includegraphics[width=0.6\columnwidth]{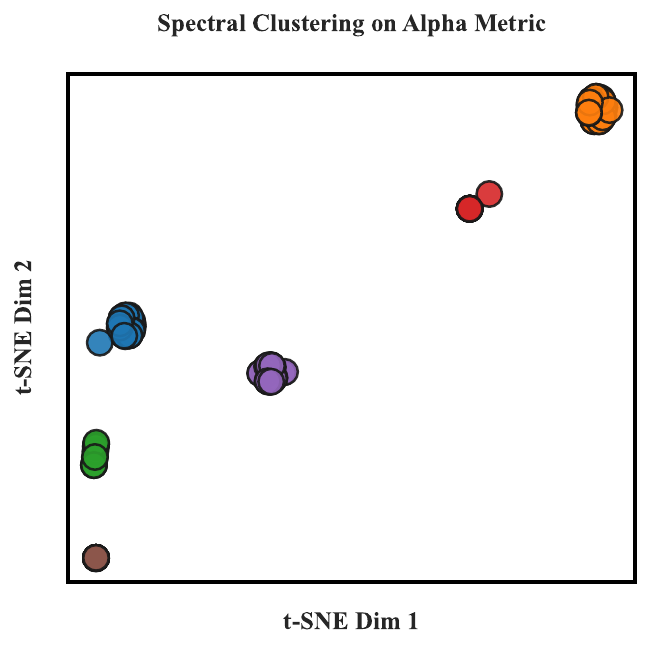}
    \includegraphics[width=0.7\columnwidth]{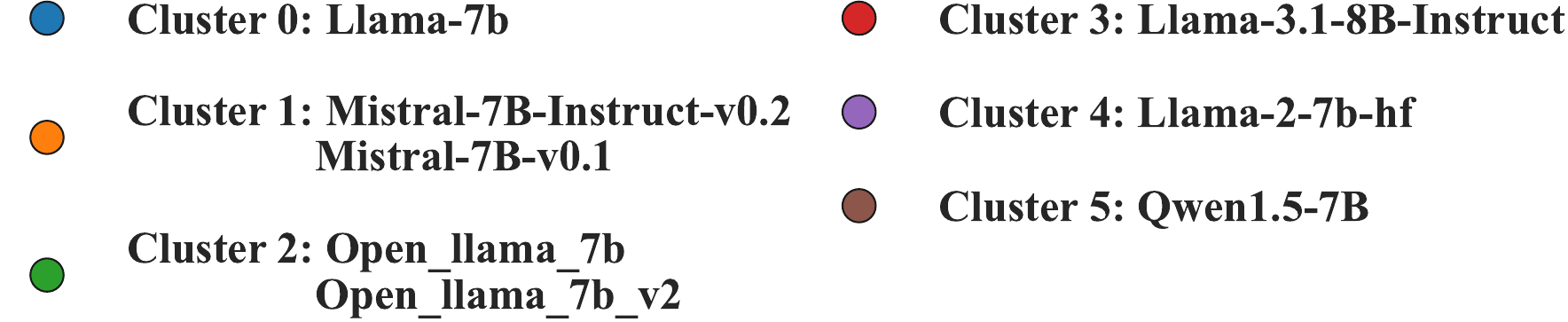}
    \caption{t-SNE 2D embedding for spectral clustering on \ALPHAHILL with ground truth cluster labels.}
    \label{fig:tsne}    
\end{figure}

In Fig.~\ref{fig:tsne}, we show a t-SNE visualization of a 2D embedding for spectral clustering on \ALPHAHILL with ground truth labels.

\subsection{Additional Generalization Tests}
\label{app:scale}

\paragraph{Model Scale Generalization.} 
As shown in Tab.~\ref{tab:scale}, we evaluate additional models at 1.3B (GPT-Neo family) and 70B (Llama-3.1 family); related pairs achieve similarity $\geq$ 0.99 while unrelated pairs score 0.55–0.69, consistent with the main results.

\begin{table}[!ht]
\caption{Similarity scores for additional model scales.}
\label{tab:scale}
\vspace{-2mm}
\centering
\resizebox{0.7\columnwidth}{!}{
\begin{tabular}{lcc}
\toprule
\textbf{Model} & \textbf{Relation} & \textbf{Similarity} \\
\midrule
Llama-3.1-70B-Instruct & Related & 0.99 \\
OpenMath2-Llama3.1-70B & Related & 0.99 \\
Qwen2.5-72B & Unrelated & 0.55 \\
\midrule
lamini-neo-1.3B & Related & 1.00 \\
lamini-cerebras-1.3B & Unrelated & 0.68 \\
Quokka-1.3B & Unrelated & 0.69 \\
\bottomrule
\vspace{-5mm}
\end{tabular}}
\end{table}

\paragraph{Intra-family Performance Prediction.} 
To further validate that spectral signatures capture within-family performance variation, we evaluate fine-tuned models with substantial benchmark gains over their base models ($>5$ points). As shown in Tab.~\ref{tab:intra}, the predicted scores track the true scores rather than collapsing to the base-model prediction. For the Llama-13B family, our method achieves MAEs of 1.94 on TruthfulQA and 1.55 on MMLU, with significant rank correlation in 7 out of 8 settings.

\begin{table}[!ht]
\caption{Intra-family performance prediction using statistically significant Kendall's Tau correlation. }
\label{tab:intra}
\centering
\resizebox{0.6\columnwidth}{!}{
\vspace{-5mm}
\begin{tabular}{lcc}
\toprule
Benchmark & Llama-2-13b & Llama-13b \\
\midrule
MMLU & $0.8347$ & $0.0004$ \\
ARC-C & $0.0272$ & $0.0002$ \\
HellaSwag & $0.0061$ & $0.0048$ \\
TruthfulQA & $0.0086$ & $0.0091$ \\
\bottomrule
\vspace{-5mm}
\end{tabular}}
\end{table}

\subsection{Additional Details on Ablation Studies}
\label{ablation_app}

\paragraph{Output-Invariant Reparameterization}

We apply scale-preserving reparameterization to consecutive linear maps while keeping the composed operator unchanged. We define a scaling factor $a = \exp(\gamma z)$ with strength $\gamma$ and $z \sim \mathcal{N}(0, 1)$. The weights are updated as $Q' = a Q, K' = K/a$ (and similarly for $V, O$).
For Permutation on Hidden Units, we generate a random permutation matrix $P$ and construct a transformed weight set: 
\(
W' = \{P W_{gate}, P W_{up}, W_{down} P^{-1}\}
\)
for each layer.
The similarity between reparameterized models and their origins remains 1 for Llama-2-7b, Qwen-1.5-7b, and Mistral-7b, suggesting robustness of our method. In contrast, PCS degrades substantially under both transformations.

\paragraph{Noise Perturbation}
For each weight matrix $W_\ell$, we sample $Z_\ell \sim \mathcal{N}(0,I)$ with the same shape and define
\begin{equation}
\Delta W_\ell = \mathrm{std}(W_\ell) Z_\ell,
\qquad
W_\ell^{(\gamma)} = W_\ell + \gamma \Delta W_\ell,
\end{equation}
where $\gamma$ controls the noise strength. We pre-generate ${\Delta W_\ell}_\ell$ once and reuse it across all $\gamma$ values for controlled comparisons.

\end{document}